\documentclass{article}

\usepackage{longtable}
\usepackage{booktabs}
\usepackage{float} 

\usepackage{tabularray}
\UseTblrLibrary{booktabs} 
\usepackage{xurl}
\usepackage{array}  
\usepackage{booktabs} 
\usepackage{graphicx} 
\usepackage[table]{xcolor} 
\usepackage{microtype}
\usepackage{ragged2e} 

\usepackage{tcolorbox}
\usepackage{xcolor}
\tcbuselibrary{breakable}
\usepackage[utf8]{inputenc}
\usepackage{tcolorbox}
\tcbuselibrary{breakable, skins}
\usepackage{xcolor}
\usepackage{listings}
\newcolumntype{L}[1]{>{\raggedright\arraybackslash}m{#1}}

\definecolor{role_ceo}{RGB}{230, 242, 255} 
\definecolor{role_cto}{RGB}{235, 247, 235} 
\definecolor{role_dev}{RGB}{255, 248, 235} 
\definecolor{result_fail}{RGB}{255, 240, 240} 
\definecolor{border_fail}{RGB}{180, 40, 40} 

\newtcolorbox{roleoutput}[3][]{
    colback=#2!30!white,   
    colframe=#2!80!black,   
    title={#3},             
    fonttitle=\bfseries,
    breakable,               
    enhanced,
    arc=3mm,               
    #1                     
}

\usepackage[utf8]{inputenc}
\usepackage{tcolorbox}
\tcbuselibrary{breakable, skins}
\usepackage{xcolor}
\usepackage{listings} 

\definecolor{role_ceo}{RGB}{230, 242, 255} 
\definecolor{role_cto}{RGB}{235, 247, 235} 
\definecolor{role_dev}{RGB}{255, 248, 235} 
\definecolor{result_pass}{RGB}{235, 250, 235} 
\definecolor{result_fail}{RGB}{255, 240, 240} 
\definecolor{border_pass}{RGB}{40, 120, 40} 
\definecolor{border_fail}{RGB}{180, 40, 40} 

\newtcolorbox{rolebox}[2][]{
    colframe=black,
    fonttitle=\bfseries,
    title={#2},
    breakable,  
    enhanced, 
    #1  
}

\usepackage{multirow}  
\usepackage{booktabs} 
\usepackage{xcolor}  
\usepackage{colortbl} 
\usepackage{array}    
\usepackage{xltabular} 
\usepackage{multirow}
\usepackage[table]{xcolor}
\usepackage{tabularx} 
\usepackage{graphicx}
\usepackage{subcaption}
\usepackage{booktabs} 
\usepackage{xltabular}

\lstdefinelanguage{json}{
    basicstyle=\rmfamily\scriptsize\linespread{0.8}\selectfont,
    numbers=left,
    numberstyle=\tiny,
    stepnumber=1,
    numbersep=5pt,       
    showstringspaces=false,
    breaklines=true,
    string=[s]{"}{"},
    comment=[l]{//},
    morecomment=[s]{/*}{*/},
    morekeywords={true,false,null}
}
\usepackage{hyperref}

\usepackage{titlesec}  

\titlespacing{\chapter}{0pt}{10pt}{10pt}

\usepackage[accepted]{icml2026}
\usepackage{enumitem}
\usepackage{amsmath}
\usepackage{amssymb}
\usepackage{mathtools}
\usepackage{amsthm}

\usepackage[capitalize,noabbrev]{cleveref}

\theoremstyle{plain}

\theoremstyle{definition}

\theoremstyle{remark}

\usepackage[textsize=tiny]{todonotes}

\icmltitlerunning{ }

\begin{document}
\pagestyle{plain}
\twocolumn[
 \begin{center}
    \bfseries \Large
    Towards a Science of Collective AI: LLM-based Multi-Agent Systems   \\ Need a Transition from Blind Trial-and-Error to Rigorous Science
\end{center}
  \vspace{10pt} 
\begin{center}

    \textbf{Jingru Fan}\textsuperscript{1,*},
    \textbf{Dewen Liu}\textsuperscript{2,*},
    \textbf{Yufan Dang}\textsuperscript{3},
    \textbf{Huatao Li}\textsuperscript{1},
    \textbf{Yuheng Wang}\textsuperscript{1},
    \textbf{Wei Liu}\textsuperscript{4},\\
    \textbf{Feiyu Duan}\textsuperscript{2},
    \textbf{Xuanwen Ding}\textsuperscript{2},
    \textbf{Shu Yao}\textsuperscript{1},
    \textbf{Lin Wu}\textsuperscript{1},
    \textbf{Ruijie Shi}\textsuperscript{3},
    \textbf{Wai-Shing Leung}\textsuperscript{3},\\
    \textbf{Yuan Cheng}\textsuperscript{1},
    \textbf{Zhongyu Wei}\textsuperscript{2},
    \textbf{Cheng Yang}\textsuperscript{5},
    \textbf{Chen Qian}\textsuperscript{1,$\dagger$},
    \textbf{Zhiyuan Liu}\textsuperscript{3,$\dagger$},
    \textbf{Maosong Sun}\textsuperscript{3}

    \vspace{10pt} 

    \textsuperscript{1}Shanghai Jiao Tong University, 
    \textsuperscript{2}Fudan University
    \textsuperscript{3}Tsinghua University, \\ 
    \textsuperscript{4}King's College London, 
    \textsuperscript{5}Peng Cheng Laboratory

    \vspace{5pt} 
    \{qianc, fanjingru510\}@sjtu.edu.cn,
    dwliu23@m.fudan.edu.cn,
    liuzy@tsinghua.edu.cn

     \vspace{10pt} 
\end{center}
\printAffiliationsAndNotice{} 

]
{ 
    \renewcommand{\thefootnote}{\fnsymbol{footnote}}
    
    \footnotetext[1]{Equal contribution.}    
    \footnotetext[2]{Corresponding authors.}
}

\begin{abstract}
Recent advancements in Large Language Models (LLMs) have greatly extended the capabilities of Multi-Agent Systems (MAS), demonstrating significant effectiveness across a wide range of complex and open-ended domains. However, despite this rapid progress, the field still relies heavily on empirical trial-and-error. It lacks a unified and principled scientific framework necessary for systematic optimization and improvement. This bottleneck stems from the ambiguity of attribution: first, the absence of a structured taxonomy of factors leaves researchers restricted to unguided adjustments; second, the lack of a unified metric fails to distinguish genuine collaboration gain from mere resource accumulation. In this paper, we advocate for a transition to design science through an integrated framework. We advocate to establish the collaboration gain metric ($\Gamma$) as the scientific standard to isolate intrinsic gains from increased budgets. Leveraging $\Gamma$, we propose a factor attribution paradigm to systematically identify collaboration-driving factors. To support this, we construct a systematic MAS factor library, structuring the design space into control-level presets and information-level dynamics. Ultimately, this framework facilitates the transition from blind experimentation to rigorous science, paving the way towards a true science of Collective AI.
\end{abstract}
\section{Introduction}

With the enhanced reasoning capabilities of Large Language Models (LLMs), powerful autonomous agents capable of goal-oriented planning and task execution have emerged ~\citep{Minaee2024, Zhou2024, wang2024survey}. By integrating internal memory and external tool-use, these agents can independently execute complex tasks, such as code generation and mathematical problem-solving ~\citep{xi2025rise, schick2023toolformer}. 
While these agents are inherently limited by their individual capability boundaries, making it difficult to maintain universal expertise across open-ended domains, inspired by the success of collective intelligence in biological swarms and human civilizations (detailed in Appendix \ref{appendix:evolution}), Multi-Agent Systems (MAS) have emerged as a promising direction to break through these constraints~\citep{chen2023agentverse,Li2024survey,Guo2024}. By organizing diverse individuals to work collaboratively, MAS can transcend some limitations of individual agents, enabling agents to solve multi-turn, multi-function, and open-ended problems through collective intelligence~\citep{tran2025multi,Maldonado2024}. Consequently, the MAS paradigm has achieved significant success in complex and open fields such as scientific research~\citep{Robin2025}, software engineering~\citep{qian2024chatdev}, infrastructure management ~\citep{Seff2023}, healthcare services~\citep{Liu2025}, financial analysis~\citep{Xiao2024}, and social sciences~\citep{Yang2024}.

\begin{figure*}[!ht]
    \centering
    \includegraphics[width=\textwidth]{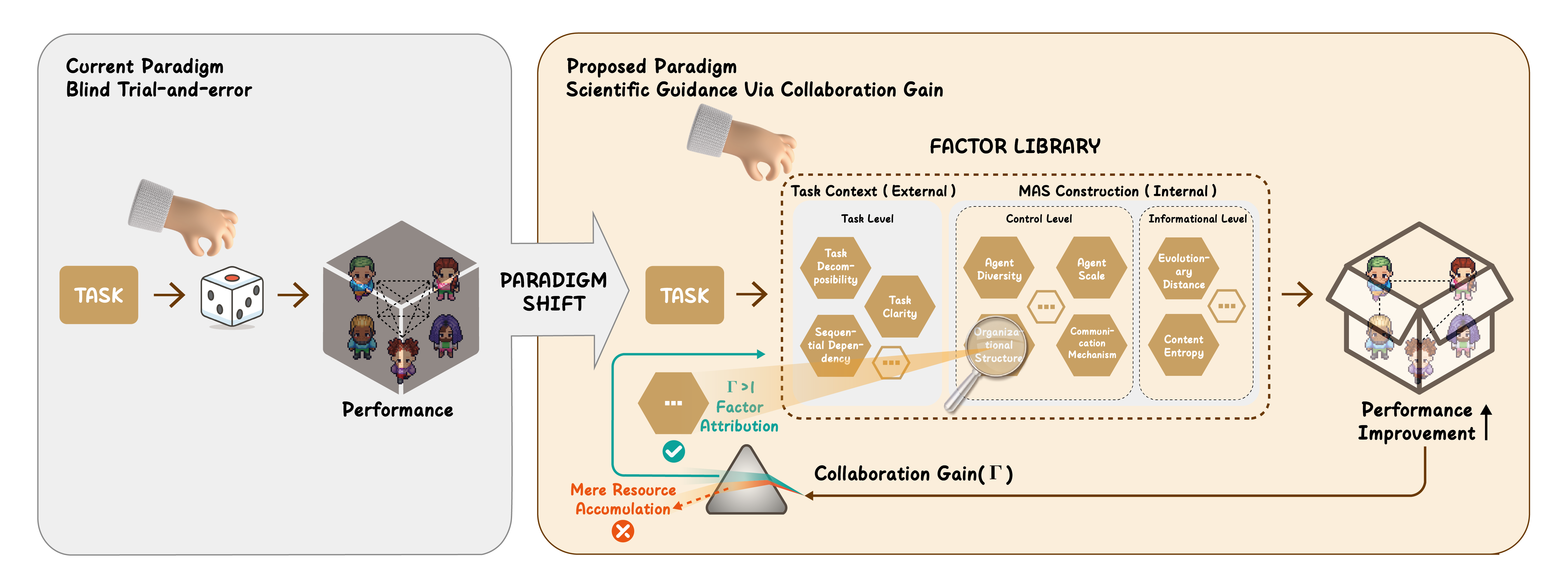}
    \caption{The Paradigm Shift: From Blind Trial-and-Error to a Science Guidance.
    Left (Current): An opaque black box where performance gains are stochastic and unattributable.
    Right (Proposed): A white-box paradigm. Researchers select factors from the library to construct the MAS; observed performance is then passed through the $\Gamma$ (the prism), which filters out mere resource accumulation to isolate genuine collaboration gain. This analytic step inherently executes factor attribution.
    }
    \label{fig:section1}
\end{figure*}

However, despite the abundance of successful multi agent systems, the field's advancement remains reliant on engineering-driven blind trial-and-error, lacking scientific guidance for systematic improvement~\citep{Cemri2025,Guo2024}. Researchers thus default to optimizing performance via empirical adjustments to key factors—such as organizational structures~\citep{Li2024survey}. This reliance on empirical adjustments essentially stems from the inability to conduct rigorous factor attribution. While such empirical optimization occasionally yields performance improvements, the true attribution of these gains remains obscured. 

From the perspective of MAS construction, the ambiguity of attribution manifests as two critical challenges. Firstly, prior to initiating MAS optimization, researchers must select specific factors to verify performance gains; however, this selection is hindered by the absence of a structured taxonomy for the vast factor space~\citep{Maldonado2024}. Without a standardized candidate factor set to anchor the process, researchers are restricted to unguided factor adjustments, unable to systematically examine potential collaboration-driving factors~\citep{kim2025towards}. Even more critically, in the dimension of value verification, the lack of a unified metric makes it impossible to distinguish whether a specific factor modification drives genuine collaboration gain or merely reflects mere resource accumulation(capacity growth without collaboration)~\citep{Zhang2025}. Current task-specific metrics inherently conflate these two qualitatively distinct effects: the intrinsic gain from collaboration (i.e., capability growth resulting from agent collaboration) and the performance improvement resulting merely from increased computational budget~\citep{qian2024scaling}. Lacking both a structured framework for factors and a precise diagnostic signal for metrics, researchers cannot systematically identify the design factors that drive genuine collaboration~\citep{renney2026}. This limitation wastes substantial computational resources on ineffective designs and, more critically, precludes the establishment of scientific principles, preventing MAS from evolving into a rigorous science~\citep{kapoor2024}.

To address these challenges and guide the scientific construction of MAS, we propose three core recommendations that form an integrated paradigm, as illustrated in \cref{fig:section1}. Recognizing that a scientifically valid evaluation standard is the prerequisite for distinguishing effective designs, we prioritize the proposal of the metric and attribution methodology, which then governs the systematic exploration of the factor space~\citep{zhu2026raffles}:

\begin{enumerate}
\item \textbf{Build Collaboration Gain Metric ($\Gamma$):}
First and foremost, we call for establishing the collaboration gain metric ($\Gamma$) as a quantified feedback signal for system optimization. We define $\Gamma$ as the performance ratio of a MAS to a Single-Agent System (SAS) under equivalent computational resource constraints. By isolating the genuine collaboration gain from mere resource accumulation~\citep{Du2024}, this metric provides the essential quantitative feedback needed to identify valid optimization directions. It thus serves as a principled diagnostic signal that anchors the entire scientific framework.

\item \textbf{Factor Attribution Paradigm:}
Leveraging this metric, we propose designing a factor attribution paradigm to identify genuine collaboration-driving factors. The core mechanism follows a two-step, sequential logic: we first examine whether modifying a factor leads to measurable performance improvement. Only when such improvement is observed do we apply the $\Gamma$ metric for scientific evaluation to verify the source of the gain—a candidate factor is confirmed as a genuine collaboration contributor if and only if $\Gamma > 1$, which indicates the improvement stems from genuine collaboration gain rather than mere resource accumulation~\citep{lamalfa2025largelanguagemodelsmiss}. This sequential validation can help transform factor attribution from trial-and-error into a more rigorous, evidence-based scientific process.

\item \textbf{Construct Factor Library:}
To support this attribution, we advocate for a systematic MAS factor library—a systematically organized set of variables influencing system behavior. While the attribution paradigm provides the ``method,'' this library provides the ``scope'' by systematizing the previously fragmented design space, providing a structured candidate factor set for the aforementioned attribution task. We decouple the factor space into the task context (external) and the MAS construction (internal) ~\citep{Maldonado2024}. The latter is further organized into two essential levels: the control level, representing static architectural presets (e.g., organizational structure)~\citep{Chen20251}, and the information level, characterizing dynamic execution mechanisms (e.g., content entropy)~\citep{Alexander2025,Zhi-qiu2025,Kuhn2023}. This structured framework provides a rigorous and reusable roadmap for systematic optimization.
\end{enumerate}
\vspace{-0.3em} 
By establishing the correlations between specific factors and collaboration gain via this integrated paradigm, we aim to replace blind trial-and-error with systematic scientific guidance~\citep{Panait2005}, a shift that can help researchers move from less systematic approaches toward more proactive modulation of MAS factors to enhance collaborative efficacy. Since the transition toward a true science of Collective AI requires more than engineering-driven blind trial-and-error, it demands a fundamental rethinking of our research trajectory and a collective consensus on scientifically guided MAS construction~\citep{zhu2026raffles}. We therefore advocate for the research community to embrace this integrated paradigm, providing a roadmap to transform MAS into a rigorous science and ensuring that the next generation of collective intelligence is built upon transparent scientific logic.
\vspace{-0.1cm}
\section{Problem Statement}
\label{sec:problem_statement}

While MAS have demonstrated outstanding performance across numerous tasks, the field remains heavily reliant on engineering-driven blind trial-and-error, unable to transition toward rigorous scientific construction~\citep{Li2024survey,Wu2025}. 
From the perspective of MAS construction, this methodology hits a bottleneck stemming fundamentally from the ambiguity of attribution. This ambiguity obscures the connection between specific factors and performance gains, severely impeding the field's evolution from random trials to systematic scientific guidance. Specifically, this manifests as two critical challenges: the unguided search among a vast array of factors, and the inability of metrics to discern genuine collaboration gain from mere resource accumulation~\citep{ma2025doverinterventiondrivenautodebugging}.

\subsection{Absence of Structured Taxonomy for Factor Selection}
\label{subsec:gap1_factors}
The first barrier lies in the difficulty of factor selection, which currently resembles a blind search due to the absence of structured scientific guidance. Since MAS optimization is governed by a vast array of interacting systemic factors—such as agent scale~\citep{qian2024scaling,Wang2025}, organizational structure~\citep{Chen20251,Zhou2025}, and communication mechanism~\citep{Zou2025,Yang2025}—navigating this high-dimensional factor space has effectively become a blind search. This unguided search becomes computationally intractable as task complexity scales. Researchers face the risk of diminishing marginal returns, as coordination overheads—including communication and synchronization—often grow faster than collaborative benefits~\citep{yang2026toward}. Recent findings even indicate that when agent scale exceeds a specific threshold, system performance tends to stagnate or even degrade due to information overload~\citep{qian2024scaling}. 

Currently, research is trapped in a double bind: we attempt to solve increasingly difficult problems by raising system complexity, yet we lack the methodology to navigate that very complexity~\citep{lamalfa2025largelanguagemodelsmiss}. This necessitates a factor-based framework to transform this blind search into proactive, principled regulation.

\subsection{Metrics Confounding Genuine Collaboration Gain and Resource Scaling}
\label{subsec:gap2_metrics}

Even more critically, distinct from the factor selection challenge, valid attribution is prevented by the second gap: the inability of existing metrics to distinguish genuine collaboration from resource accumulation. The persistent reliance on blind trial-and-error is primarily driven by the fact that current evaluation methodologies tend to depend on metrics confined to task-specific performance~\citep{sun2025collabovercooked,Schipper2025,Li-qiang2024,White2025,Chan2023}. While effective for benchmarking end-results, these metrics inherently conflate two qualitatively distinct effects: the intrinsic gain from collaboration (i.e., capability growth resulting from agent collaboration) and the performance improvement resulting merely from increased computational budget~\citep{qian2024scaling}. 

Consequently, it remains difficult to precisely determine whether a MAS architecture has truly achieved an expansion of the system's capability boundaries via collective intelligence, or has merely exploited a greater volume of resources—such as increased token consumption or agent scale—compared to a single-agent baseline~\citep{Du2024}. This ambiguity obscures the causal link between systemic factors and performance outcomes, hindering the field's transition toward more predictive scientific guidance~\citep{Kuhn2010}.

\section{Measuring Genuine Collaboration Gain: A Principled Metric to Guide Factor Attribution}
\label{sec:genuine-collaboration-gain}
To resolve this ambiguity of attribution and move beyond blind trial-and-error, we advocate for the construction of a diagnostic signal that quantifies a system's genuine collaboration gain as a rigorous feedback mechanism for optimization~\citep{bo2024reflective}. While raw resource accumulation—such as expanding token budgets or agent populations—can indeed enhance performance, the diminishing marginal utility inherent in single-agent scaling suggests a fundamental ceiling for individual. Our proposed metric, therefore, serves as more than a performance benchmark; it functions as a rigorous filter for factor attribution. By decoupling genuine collaboration gain from resource-driven improvements, this metric facilitates a more systematic identification of factors that contribute to collective intelligence, thereby supporting a more informed approach to MAS design~\citep{wang2025framework}.

\subsection{Formal Definition and Theoretical Foundations}
\label{sec:collab-gain-def}
We contend that a key objective of MAS research is to capture the nonlinear leap facilitated by collective collaboration~\citep{qian2024scaling}. This pursuit is fundamentally rooted in emergence theory, which posits that the macro-level behavior of a complex system should transcend the simple summation of its micro-level constituents~\citep{fromm2005types}. In the context of MAS, this macro-level behavior manifests as the collaboration gain derived from inter-agent collaboration, while the micro-level summation corresponds to the cumulative performance of agents acting in isolation. While we do not deny the fundamental role of individual scaling and resource expansion in driving intelligence, we argue that the true scientific frontier lies in identifying whether—and to what extent—group-level organization can yield additional dividends beyond these scaling effects. To move toward a more rigorous science of Collective AI, we need to better disentangle these emergent effects from the contributions of mere resource expansion~\citep{Hoel2016}. 

To solve this, we propose the collaboration gain ($\Gamma$), a metric that quantifies the efficiency of a collective relative to a single-agent baseline under identical resource consumption~\citep{tang2025on}. By benchmarking performance against this resource-equivalent reference, $\Gamma$ can serve as a probe for measuring genuine collaboration gain, helping to isolate the dividends of interaction from the impact of computational scale~\citep{chen2024reconcile}.

\vspace{-0.5em}
\begin{equation}
\Gamma = \frac{\Phi_{M}}{\Phi_{S}}
\footnote{We use the ratio here for its intuitive simplicity, but a more generalized form could be any function $f(\Phi_M, \Phi_S)$ that quantifies the relationship, potentially learned through regression or other methods to capture more complex, non-linear interactions.}
\end{equation}

\vspace{-1em}
\begin{itemize}
\setlength{\itemsep}{0.2em}
\item $\Phi_{M}$ represents the collective performance of the MAS, specifically capturing the efficacy facilitated by its group-level organizational and collaborative mechanisms~\citep{lamalfa2025largelanguagemodelsmiss}.

\item $\Phi_{S}$ denotes the non-collaborative baseline performance, representing the best achievable results of a single agent when allocated a total resource budget equivalent to that of the MAS.
\end{itemize}
\vspace{-1em}

Under the null hypothesis of zero interaction effect, the system performance would simply equal the single-agent baseline ($\Phi_m = \Phi_s$), yielding $\Gamma = 1$~\citep{Zhang2025}. Consequently, $\Gamma = 1$ serves as the theoretical floor for collaboration; any value above this threshold ($\Gamma > 1$) suggests that the system has achieved synergy, where ``the whole is greater than the sum of its parts''(see \cref{fig:section3}).

To operationalize this metric, we emphasize that the operational instantiation of both $\Phi_{\cdot}$ and the definition of resource equivalence are task-dependent~\citep{zhu2025multiagentbench}. First, the evaluation function $\Phi$ is adaptively selected to reflect the intrinsic goals of the task (e.g., accuracy, coverage, or efficiency)~\citep{mohammadi2025evaluation}. Second, the best achievable performance for $\Phi_{S}$ implies a saturated baseline tailored to the task's logical structure, ensuring the single agent utilizes the most effective non-collaborative strategies within the budget(detailed in Appendix ~\ref{app:baseline-std}). Finally, the metrics for resource consumption—whether measured in token throughput, sampling steps, or tool-invocation counts—are calibrated to the specific bottlenecks of each task (detailed in Appendix ~\ref{app:Resource Equivalence}). 

\subsection{$\Gamma$-Driven Analysis of Factor Attribution}
\label{sec:binary-protocol}
Building upon this foundation, $\Gamma$ functions as more than a performance metric; it serves as a diagnostic probe for isolating the actual collaborative yield of collective intelligence. However, simply calculating a number is insufficient for meaningful system design. Given the chaotic landscape of MAS design, attempting quantitative modeling (e.g., regression) on unverified factors introduces significant noise. Therefore, we establish a binary attribution framework to filter candidate factors into two distinct regimes based on their empirical validity.

\subsubsection{Categorization of Collaboration Gain}
We use $\Gamma = 1$ as the definitive boundary of synergetic emergence—the equivalence point where architectural gain meets resource cost~\citep{tang2025on}. This allows us to partition the factor space into two distinct factor categories:

\begin{figure}[t]
    \centering
    \includegraphics[width=1\linewidth,height=0.85\linewidth]{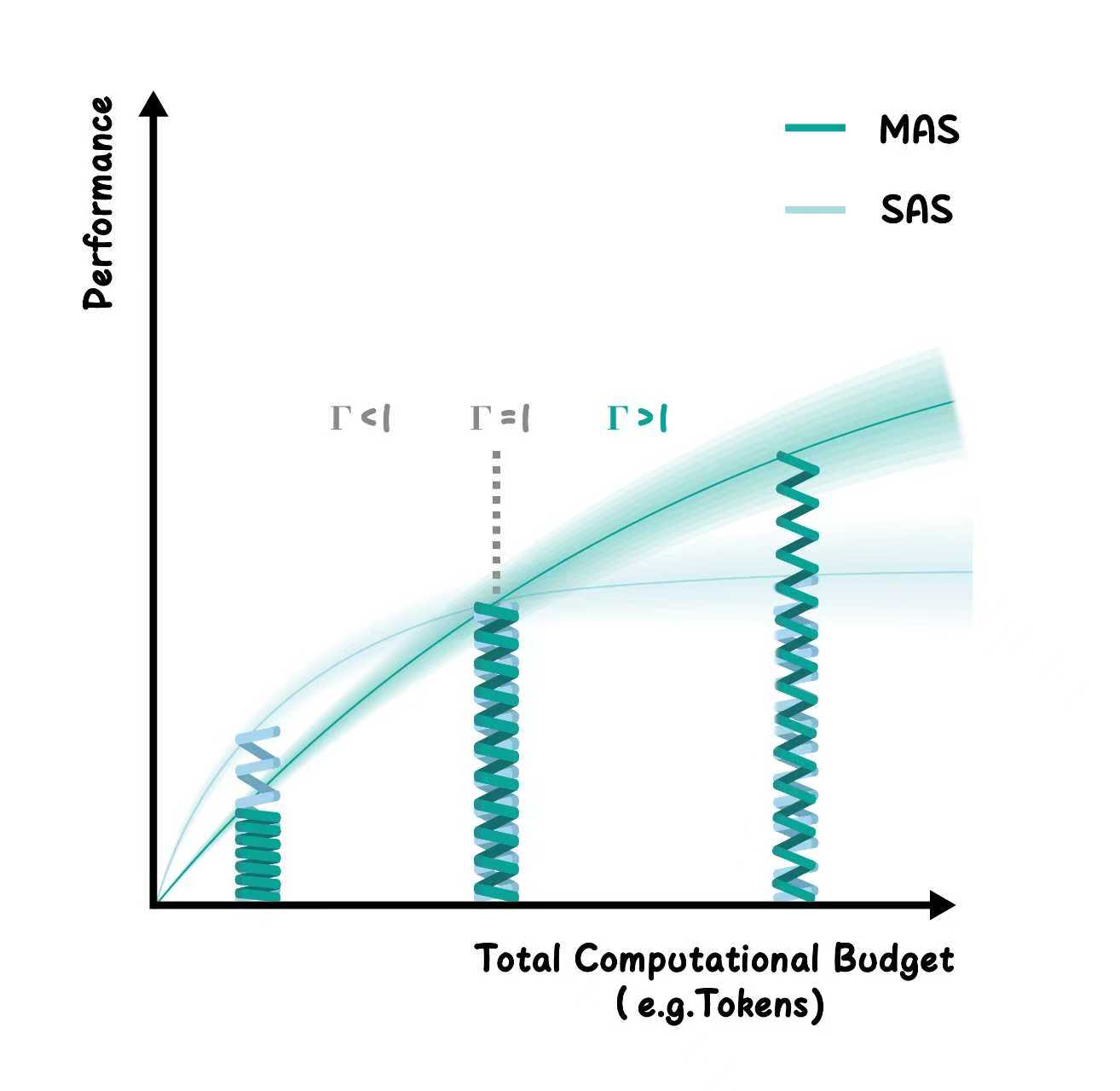}
    \caption{Conceptual Illustration of Collaboration Gain ($\Gamma$). The curves represent the performance of MAS and SAS under equivalent computational budgets to ensure comparability. When SAS performance equals or exceeds MAS, $\Gamma \lesssim 1$, indicating mere resource accumulation; conversely, $\Gamma > 1$ signifies the genuine collaboration beyond the single-agent.}
    \label{fig:section3}
\end{figure}
\vspace{-0.5em} 
\paragraph{Class I: The Positive Factors ($\Gamma > 1$)}
Factors yielding $\Gamma > 1$ with statistical significance represent the successful realization of collaboration gain. This classification confirms that the specific design (e.g., a specific communication topology) enhances system performance beyond the single-agent limit under identical resource constraints~\citep{kim2025towards}. These factors are considered true collaboration drivers, as they demonstrate a proven ability to push the system beyond individual capability ceilings. They are the valid candidates for further quantitative refinement and tuning.
\vspace{-0.5em} 
\paragraph{Class II: The Negative Factors ($\Gamma \lesssim 1$)}
Factors that fail to significantly outperform the baseline are classified into the negative set, indicating a state of attribution failure where the MAS collaboration lacks a clear advantage. This regime, characterized by $\Gamma \lesssim 1$, represents the absence of a synergetic dividend. In this state, any potential benefits of collaboration are either nullified by structural redundancy—where the protocol merely replicates individual reasoning—or actively suppressed by negative interference, such as coordination overhead and context fragmentation~\citep{Cemri2025, Zhang2025}. Consequently, the negative factors serve as a critical pruning signal, compelling researchers to recognize that the current architecture fails to justify its added complexity and preventing the wasteful optimization of ineffective designs.

\subsubsection{The Factor Attribution Process} 
To operationalize this framework for factor attribution, we advocate for a sequence that begins by strictly preconditioning the experiment on a fixed computational budget (e.g., total token consumption)~\citep{kim2025towards}. This budget is utilized to construct a saturated single-agent baseline $\Phi_S$, ensuring it reflects the maximal non-collaborative capability within the allocated constraints, and to measure the MAS performance $\Phi_M$ under this identical resource allocation. We then utilize $\Gamma$ to measure the genuine collaboration gain, effectively decoupling structural advantages from mere resource expansion~\citep{Li2024survey}. To verify the validity of the factor attribution against stochastic noise, we implement a stability filtering stage; a factor is promoted to the positive set if it maintains a sustained $\Gamma > 1$ advantage. Finally, this process culminates in the final attribution, identifying whether the specific architectural design functions as a verified causal driver of the observed collaboration gain. This approach ensures that any MAS optimization is grounded in verified collaboration gain, preventing the wasteful fine-tuning of ineffective structures(detailed in Appendix~\ref{appendix:feasibility_validation}).

\section{Systematizing the MAS Design Space: A Structured Factor Library}
\label{sec:factor-library}
\begin{figure*}[t]
    \centering
    \includegraphics[width=1\textwidth]{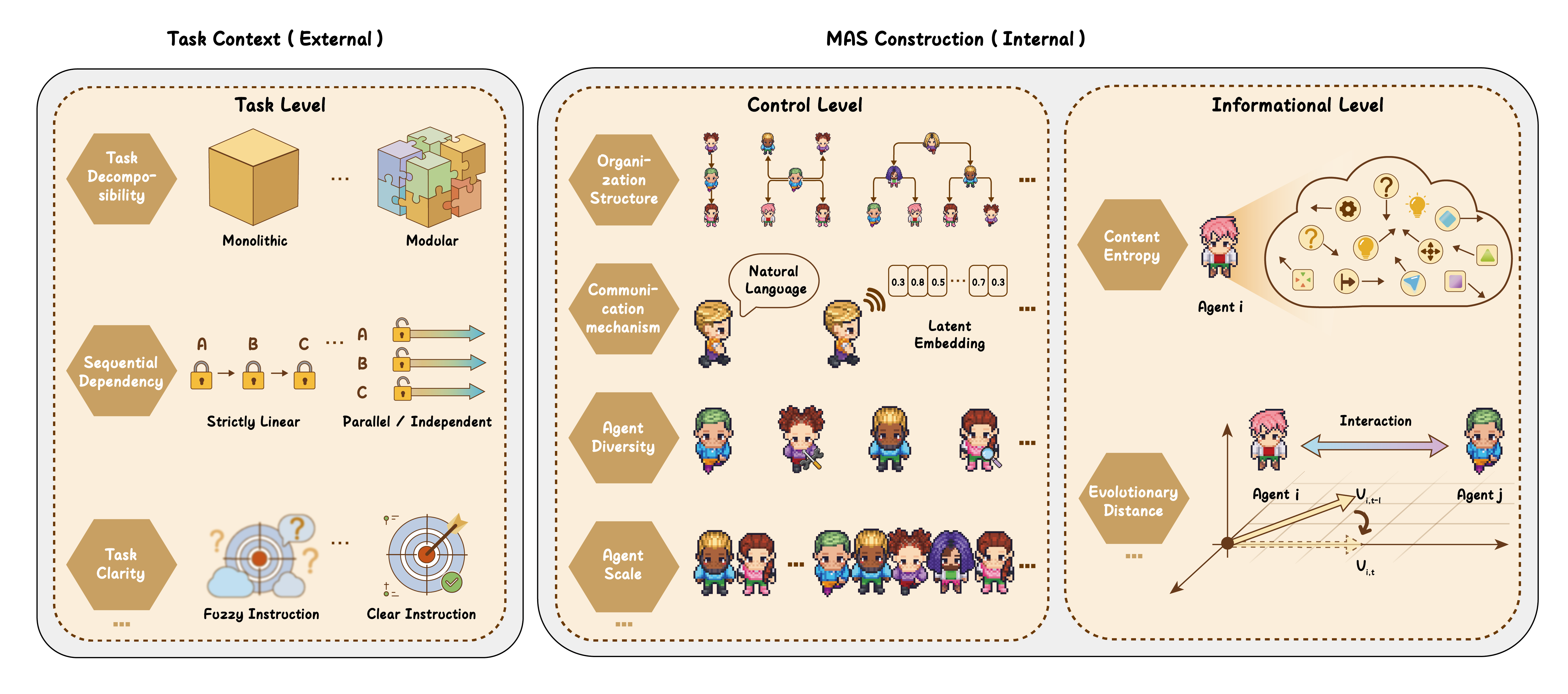}
    \caption{The MAS Factor Library Taxonomy. Factors are organized into task context(external) and MAS construction( internal), with the internal dimension spanning static control level presets and dynamic information level to guide rigorous system design.}
    \label{fig:section4}
\end{figure*}

To facilitate effective factor attribution, we propose constructing a MAS factor library(see \cref{fig:section4}). This library systematizes the previously fragmented design space of MAS factors, providing a structured candidate factor set for subsequent factor attribution.

This framework enables the factor attribution process to transcend reliance on blind trial-and-error, instead facilitating scientific exploration through a rigorous taxonomy comprising two primary domains: the task context(external) and the MAS construction(internal). Within the internal construction, we further stratify factors into two essential levels: the control level, representing static architectural presets (e.g., organizational structure), and the information level, characterizing dynamic execution mechanisms (e.g., content entropy)~\citep{Li2024survey,Maldonado2024}. This classification provides a rigorous roadmap, ensuring that the construction of MAS is guided by a comprehensive and logical design space. For specific optimization strategies and literature-supported implementations corresponding to each factor, please refer to Appendix \ref{appendix:factor_library}.

\subsection{Task Context (External) Factors}

The attributes of a task define the external boundary conditions under which a MAS operates~\citep{Decker1995}.They shape the logical structure of the problem space and constrain the bounds of the collaboration gain $\Gamma$. In particular, task attributes such as task decomposability, sequential dependency, and task clarity provide representative illustrations of how task context can affect the compatibility between problem formulation and collaborative agent architectures~\citep{Davis1983Negotiation,Malone1994,Decker1993}.

Empirical evidence suggests that mismatches between task structure and MAS design can neutralize or even degrade collaborative benefits~\citep{kim2025towards}. For example, highly sequential tasks with strong dependency chains often limit parallelism, reducing the effectiveness of distributed coordination, while tasks with low inherent clarity often necessitate explicit task decomposition and planning structures (e.g., in software engineering)~\citep{qian2024chatdev}.

These properties appear to shape the feedback behavior of $\Gamma$, suggesting that the effectiveness of collective cooperation is inherently influenced by the task's execution logic.

\subsection{MAS Construction (Internal) Factors}
We systematize the MAS factor space into two collaborative levels: control and information level. The control level encompasses static architectural presets, serving as the foundation that limits collaboration potential~\citep{Li2024survey,Guo2024,Maldonado2024}. Complementarily, the information level characterizes dynamic execution mechanisms, revealing how this potential is activated and transformed into actual gains~\citep{Parunak2001Entropy,Nowak2006Evolutionary}.

\subsubsection{Control Level}

Control level encompasses static architectural factors, serving as the foundation that presets collaboration potential limits~\citep{Li2024survey}. Its core logic formally maps unstructured requirements into a structured system design, defining operational boundaries via specific structural factors. This level integrates four dimensions: organizational structure, communication mechanisms, agent diversity, and agent scale~\citep{Guo2024,Maldonado2024}. Improper design creates structural bottlenecks, resulting in poor collaboration gain.

\textbf{Organizational structure} constitutes the architectural configuration and connection topology of the MAS, establishing the fundamental pathways available for agent interaction~\citep{Li2024survey}. We identify this as a critical factor because the potential of even high-performing agents can be constrained by a restrictive topology, which may impose bottlenecks that limit information propagation and collaborative efficiency.~\citep{qian2024scaling}.

Research in this domain is undergoing a transition from preset, static hierarchies to adaptive, dynamic architectures. For example, agentic supernets allow systems to replace fixed workflows with flexible topologies, achieving higher accuracy at only a fraction of traditional inference costs~\citep{Zhang2025}. A well-constructed topology is designed to optimize the density of connections and minimize structural redundancy, thereby providing the architectural foundation necessary for the collaboration gain to manifest as tangible performance improvements.

\textbf{Communication mechanism} governs internal collaboration and environmental feedback, encompassing modalities ranging from explicit natural language to implicit latent space representations~\citep{li2023camel}. Particularly within text-based interactions, unconstrained protocols often lead to a ``communication explosion'', where redundant dialogues cause exponential resource scaling and diminishing marginal returns~\citep{kim2025towards,wang2024survey,tran2025multi}. We argue that excessive and unstructured redundancy can suppress $\Gamma$ by drowning critical task signals in communicative noise, thereby preventing the system from exceeding single-agent performance under the same budget.

To address this issue, recent research focuses on communication-efficient design, optimizing interaction structures and protocols to reduce redundancy while preserving coordination capacity~\citep{Chen2025}. 
For example, one-shot pruning on spatial-temporal message graphs filters redundant transmissions, reducing token usage by 28.1\%--72.8\% while maintaining system robustness~\citep{Zhang2024}. Through such structured coordination control, the mechanism lowers coordination costs and seeks to create the necessary conditions for converting structural potential into a verified collaboration gain ($\Gamma > 1$). 

\textbf{Agent diversity} characterizes the degree of functional heterogeneity within a MAS, representing a tunable factor of differentiation. Formally, let $\mathcal{I} = \{i_1, \dots, i_N\}$ be the set of agents, where each agent $i_k$ is defined by the tuple $\mathcal{A}_k = \langle P_k, M_k, T_k, R_k \rangle$, denoting base parameters, memory mechanisms, toolsets, and role configurations, respectively ~\citep{wang2024survey,xi2025rise,schick2023toolformer}. Instead of a fixed state, diversity is a range that can be smoothly adjusted; by changing how much these functional parts overlap or differ, we can control whether the system stays uniform or becomes highly specialized.

We argue that functional heterogeneity acts as a compensatory mechanism for individual cognitive bias, expanding the collective solution space while enabling cross-verification~\citep{Dongfu2023}. By carefully adjusting differences in toolsets ($T_k$), roles ($R_k$), memory structures ($M_k$), or reasoning modalities, one can strategically amplify complementary strengths among agents, thereby boosting collective performance~\citep{Rui2025}. By synthesizing specialized perspectives, diversity provides a potential pathway to realize a positive collaboration gain ($\Gamma > 1$).

\textbf{Agent scale} defines the total number of agents, which forms the basis for interaction complexity. As scale increases, the expanded solution space and interaction paths provide a structural foundation for collaborative emergence. Specifically, coordinating over 1,000 agents facilitates a logistic growth pattern, allowing collaborative emergence to appear much earlier than in traditional neural scaling~\citep{qian2024scaling}. For instance, the OASIS system supports millions of agents, demonstrating how larger scales can simulate stronger collective dynamics~\citep{Yang2024}. Similarly, in financial trading TradingAgents achieves better performance through multi-role collaboration~\citep{Xiao2024}. In this framework, agent scale serves as a tunable parameter that opens a pathway for $\Gamma$ to manifest as significant collective intelligence through phase transitions in performance.

\subsubsection{Information Level}

Information level characterizes the dynamic activation of potential into actual gains during execution. While control factors define the architectural boundaries, information factors—emerging in real time—serve as proxies for the system's internal cognitive trajectory. We include this level to provide mechanistic transparency into the collaborative process, enabling quantitative tracking of how static configurations manifest as dynamic collaboration gain $\Gamma$. We analyze two dimensions: \textit{intra-agent} uses content entropy to measure solution certainty and convergence, while \textit{inter-agent} uses evolutionary distance to track semantic flow and interaction contributions.

\textbf{Content entropy} measures the certainty of the solution space by capturing intent distribution within the semantic feature space~\citep{Zhi-qiu2025}. The instantaneous content entropy at time $t$ is defined as:

\vspace{-0.5em} 
\begin{equation}
H_t = -\sum_{i} p(x_i | C_t) \log p(x_i | C_t)
\end{equation}
\vspace{-0.5em} 

where $H_t$ is the entropy at time $t$, $i$ indexes the set of possible discrete content types generated by agents, and $p(x_i \mid C_t)$ is the conditional probability of content $x_i$ given system state $C_t$.

In the MAS framework, content entropy $H_t$ serves as a quantitative proxy for the system's informational state transition. A monotonic decrease in $H_t$ characterizes the convergence trajectory from stochastic exploration toward a stabilized collective state~\citep{Navajas2022}. Conversely, persistent high entropy signifies a coordination deficit, where the lack of effective constraints leaves the system in a state of decision-making divergence ~\citep{Guo2025}. However, since a reduction in entropy only describes agreement in a formal sense, it does not necessarily equate to a high-quality solution~\citep{Yang2025}. For instance, in cases of ``contextual breakdown,'' the system may exhibit pseudo-convergence because agents prematurely ignore key information. Therefore, a simple decrease in entropy cannot distinguish between valid consensus and degenerative information loss, requiring a deeper analysis of the interaction context, with more detailed examples provided in Appendix~\ref{appendix:feasibility_validation}.

\textbf{Evolutionary distance} characterizes the dynamic ``work'' of the system, quantifying the intensity of semantic displacement during the interaction process. This factor defines the degree to which interaction behaviors induce changes in the internal states of agents from an information-theoretic perspective ~\citep{Jaques2022}:

\vspace{-0.5em} 
\begin{equation}
D_t = \sum_{i=1}^{N} \left( 1 - \frac{\mathbf{v}_{i,t} \cdot \mathbf{v}_{i,t-1}}{\|\mathbf{v}_{i,t}\| \|\mathbf{v}_{i,t-1}\|} \right)
\end{equation}
\vspace{-0.5em} 

where $D_t$ denotes the evolutionary distance of the system at time $t$, characterizing the overall update intensity of semantic content relative to the previous round. Here, $N$ represents the total number of agents in the MAS, and $i$ is the agent index. The term $\frac{\mathbf{v}_{i,t} \cdot \mathbf{v}_{i,t-1}}{\|\mathbf{v}_{i,t}\| \|\mathbf{v}_{i,t-1}\|}$ measures the cosine similarity between the state vectors(e.g., semantic state embeddings) of the $i$-th agent at consecutive time steps; subtracting this value from $1$ gives the cosine distance, which reflects the change in the agent's semantic state.

In an ideal execution, an optimal evolutionary distance suggests the emergence of meaningful new information~\citep{cover1999elements}. However, abnormal fluctuations in this factor require a two-sided analysis: a change that is too small typically indicates that the system is stuck in redundant repetition. Conversely, an excessive change may not be a positive signal, as it often suggests a contextual breakdown, where agents may lose their connection to previous context, leading to outputs completely decoupled from the preceding logic. Therefore, only when evolutionary distance and content entropy maintain a balanced relationship can the system achieve effective performance growth while staying logically consistent~\citep{march1991exploration}. An empirical analysis with detailed case studies is provided in Appendix~\ref{appendix:feasibility_validation}.

\section{Alternative Views}
This paper proposes steering MAS design onto a structured, scientific track. While founded on robust logic, this approach may encounter skepticism. This section addresses three primary counterarguments.

\subsection{Operational Complexity and Resource Costs}
Critics argue that our framework is too difficult to implement in practice. They point out that analyzing so many different factors creates an overwhelming amount of work, and that measuring the specific collaboration gain requires experimental setups that are far too strict~\citep{Dorri2018}. From this perspective, $\Gamma$ is seen as an unnecessary burden; critics prefer simpler, more direct metrics like absolute accuracy, which are much easier to track~\citep{Windl2022}.

However, we argue that prioritizing short-term convenience over understanding why a system actually works creates a significant blind spot. Simple metrics fail to reveal whether performance stems from mere resource scaling or from genuine architectural synergy. Instead of continuing with blind trial-and-error, the field should move toward adopting collaboration gain as a precise feedback signal to evaluate the true benefits of specific design factors~\citep{Fang2011}. While this rigorous evaluation requires more controlled experimentation, it provides the necessary transparency to distinguish between superficial improvements and fundamental architectural breakthroughs. This shift is essential for transforming MAS construction from an empirical craft into a predictable and reproducible science~\citep{Sharma2025}.

\subsection{Contradiction between Holism and Reductionism}
Critics from systems theory argue that breaking a MAS down into individual factors is too simplistic. They worry that by focusing on separate parts, we lose sight of ``emergence''—the way a system’s collective behavior becomes greater than the sum of its parts~\citep{Mazzocchi2012,Pigliucci2014}.

In response, we argue that systematic analysis is a necessary step toward understanding collective power. Just as we study individual genes to comprehend the complexity of life, we must identify specific design factors to see how they interact to create emergence~\citep{Fang2011}. Rather than ignoring the ``whole'', we advocate for a perspective that provides a clearer roadmap to see how these individual ``ingredients'' combine to produce collective intelligence. By identifying the role of each factor, we move from simply observing that a system works to understanding how its internal components drive that success~\citep{Carrodano2025}.

\subsection{Ambiguities in Transitioning from Correlation to Causal Discovery}
Skeptics suggest that identifying design factors yields only statistical correlations rather than true causal mechanisms, risking a static repository that fails to generalize to new environments~\citep{Bai2022,Shu2023}.

In response, we argue that factor attribution is a necessary foundation for deeper understanding. We call upon the community to adopt $\Gamma$ as a standard feedback signal to reflect the true efficacy of collective collaboration. By using this signal, we can systematically identify which design factors actually drive performance gains. This process provides the essential data needed for future causal modeling—a methodology that maps the direct link between specific design ``actions'' and their resulting ``effects''. Establishing this foundation will pave the way toward a more transparent and designable framework, shifting the field from merely observing that a system works to predicting how to make it work, facilitating that collective intelligence becomes a reproducible outcome~\citep{Scholkopf2019}.

\section{Conclusion}
Challenging the prevailing reliance on engineering-driven blind trial-and-error, this paper advocates for a paradigm shift to resolve the fundamental ambiguity of attribution in MAS. We propose an integrated framework centered on three core strategic recommendations. First, we advocate for adopting the collaboration gain metric as a scientific evaluation tool to decouple genuine collaboration gain from mere resource accumulation. Building upon this metric, we propose a factor attribution paradigm to systematically identify the true causal drivers of performance. Finally, we call for constructing a systematic MAS factor library to provide a structured design space for these attribution tasks. Collectively, these recommendations aim to transition the construction of MAS from empirical practices toward systematic scientific guidance. By ensuring that performance improvements are theoretically traceable, we offer a roadmap for the field to evolve into a rigorous science, facilitating the reliable engineering of collective intelligence.

\bibliography{example_paper}
\bibliographystyle{icml2026}

\newpage
\onecolumn
\appendix

\section*{Appendix}
\label{sec:appendix_toc}

\section{Theoretical Background of Collective Intelligence}
\label{appendix:evolution}
\subsection{Collective Intelligence in Biological Systems}
\label{sec:swarm_intelligence_in_bio}
The concept of \textit{Swarm Intelligence} in artificial intelligence~\cite{eberhart2001swarm} originates from the observation of cooperative behaviors in biological populations~\cite{bonabeau1999swarm}. Cooperative swarm behaviors of organisms can typically be categorized into three paradigms.

\begin{itemize} \item \textbf{Decentralized Swarm Intelligence.} In this paradigm, there is no leader or centralized controller within the system. Instead, the complexity and functionality of swarm behavior arise entirely from local interactions among individuals. The macroscopic performance of the collective far exceeds the sum of individual capabilities. For example, army ants exhibit highly organized ``raiding'' formations during foraging and can dynamically construct living bridges made of tens of thousands of ants to cross obstacles~\cite{franks1989army}. Another iconic example is the murmuration of starlings, where thousands of birds perform synchronous, fluid, and instantaneous aerial maneuvers to evade predators. Studies show that such large-scale coordination occurs without a leader; each bird only needs to align its speed and direction with those of its 6–7 nearest neighbors~\cite{ballerini2008interaction}. These local rules propagate through the network, allowing perturbations to travel across the entire flock~\cite{cavagna2010scale}, resulting in sophisticated collective behaviors that far surpass the intelligence of individual birds.

\item \textbf{Centralized Organization.} In this paradigm, one or a few core individuals exert decisive, top-down influence on the structure and function of the collective through their physiological state or behavior. A representative case is the naked mole-rat colony, in which a single breeding female (the queen) dominates reproduction. She releases specific chemical signals and engages in aggressive behaviors that physiologically suppress reproduction in all other females, forcing them to serve as workers for the colony~\cite{sherman2017biology}. Similarly, gorilla groups are led by a dominant male known as the ``Silverback.'' Although he does not micromanage individual activities, he holds ultimate authority over decisions such as when the group moves, rests, or changes routes, and he mediates internal conflicts while protecting the group from external threats~\cite{hoff2002mountain}.

\item \textbf{Hybrid Organization.} This paradigm combines centralized and decentralized elements. While the group has a core individual exerting special influence, many key decisions are still made collectively in a distributed manner. The European honeybee colony exemplifies this hybrid structure. The queen serves as the sole reproductive individual and maintains social cohesion through pheromones—a hallmark of centralization~\cite{seeley2009wisdom}. However, when the colony must find a new nest site, the decision-making process becomes highly democratic. Scout bees ``debate'' candidate locations using the symbolic waggle dance. The final decision is not made by the queen, but through consensus once the strength of dance signals for one location reaches a quorum~\cite{seeley2011honeybee}.
\end{itemize}

\begin{figure*}[!ht]
    \centering
    \includegraphics[width=1\linewidth]{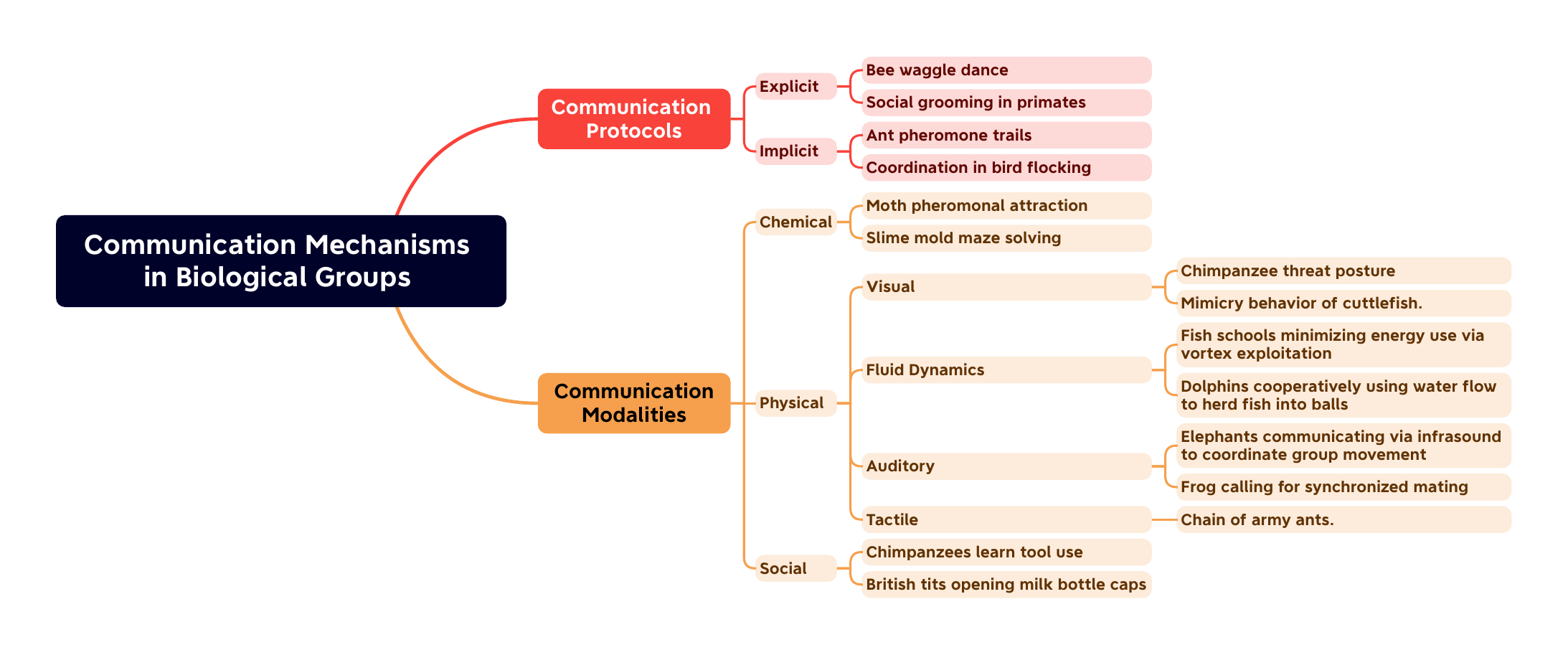}
    \caption{The taxonomy of communication mechanisms in biological groups.}
    \label{fig:sec_3.1_bio_communication}
\end{figure*}

Although individual organisms generally possess limited intelligence, evolution has endowed them with an extraordinary diversity of communication mechanisms, as illustrated in the Figure~\ref{fig:sec_3.1_bio_communication}. These mechanisms span chemical and physical modalities and even forms of social learning analogous to human behavior, transmitting rich information either explicitly or implicitly. 

In decentralized systems, numerous individuals following simple rules can self-organize into complex and adaptive macroscopic patterns~\cite{isaeva2012self}, giving rise to emergent intelligent behavior. Such phenomena have inspired a wide range of computational algorithms~\cite{sumpter2010collective}. For instance, implicit communication protocols in nature often rely on stigmergy: individuals release chemical cues that modify the environment, and others interpret these environmental changes to coordinate indirectly. This mechanism inspired multi-agent control systems and ant colony optimization algorithms. The Table~\ref{tab:sec_3.1_bio_inspired_algorithm} summarizes representative algorithms inspired by cooperative behaviors observed in biological groups.
\begin{table*}[htbp]
\centering
\small
\caption{Representative algorithms inspired by behaviors in biological groups}
\label{tab:sec_3.1_bio_inspired_algorithm}
\resizebox{\textwidth}{!}{
\begin{tabular}{l l l}
\toprule
\rowcolor{gray!10} 
\textbf{Application Domain} & \textbf{Representative Algorithm} & \textbf{Biological Inspiration} \\
\midrule
\multirow{4}{*}{Search and Optimization} 
& Ant Colony Optimization (ACO)~\cite{dorigo1996} & Ants finding the shortest path via pheromone trails \\
& Particle Swarm Optimization (PSO)~\cite{kennedy1995particle} & Social search behavior of bird flocks and fish schools \\
& Artificial Bee Colony (ABC)~\cite{karaboga2007} & Intelligent foraging and information sharing in honeybee colonies \\
& Grey Wolf Optimizer (GWO)~\cite{mirjalili2014grey} & Social hierarchy and cooperative hunting in grey wolves \\
\midrule
\multirow{3}{*}{Robotics Control} 
& Boids Algorithm~\cite{reynolds1987flocks} & Collective flight of bird flocks \\
& BEECLUST Algorithm~\cite{kernbach2009} & Aggregation behavior of young honeybees in warm hive regions \\
& Virtual Pheromone Algorithm~\cite{hoff2013} & Pheromone trail following in ant foraging \\
\midrule
\multirow{4}{*}{Bioinformatics} 
& ACO for Protein Folding~\cite{hu2008} & Ants finding shortest paths via pheromone trails \\
& ABC for Motif Discovery~\cite{karaboga2016} & Intelligent foraging behavior of bee colonies \\
& Ant-Based Clustering~\cite{lumer1994} & Ant clustering of corpses and larvae \\
\midrule
\multirow{3}{*}{Computer Networks} 
& AntNet~\cite{dicaro2018} & Indirect communication in ant foraging \\
& AntHocNet~\cite{dicaro2005} & Ant-based path discovery and maintenance \\
& Honeybee Behavior Load Balancing~\cite{ld2013} & Foraging and information-sharing behavior of honeybees \\
\midrule
\multirow{2}{*}{Social Networks} 
& PSO as a Social Model~\cite{liu2012novel} & Collective motion and information sharing in bird and fish groups \\
& ACO for Community Detection~\cite{soleimani2013finding} & Path construction via pheromone trails in ants \\
\bottomrule
\end{tabular}
}
\end{table*}

\subsection{Collective Intelligence in Human Societies}
\label{sec:human_collective_intelligence}

\begin{figure*}[t]
    \centering
    \includegraphics[width=1\linewidth]{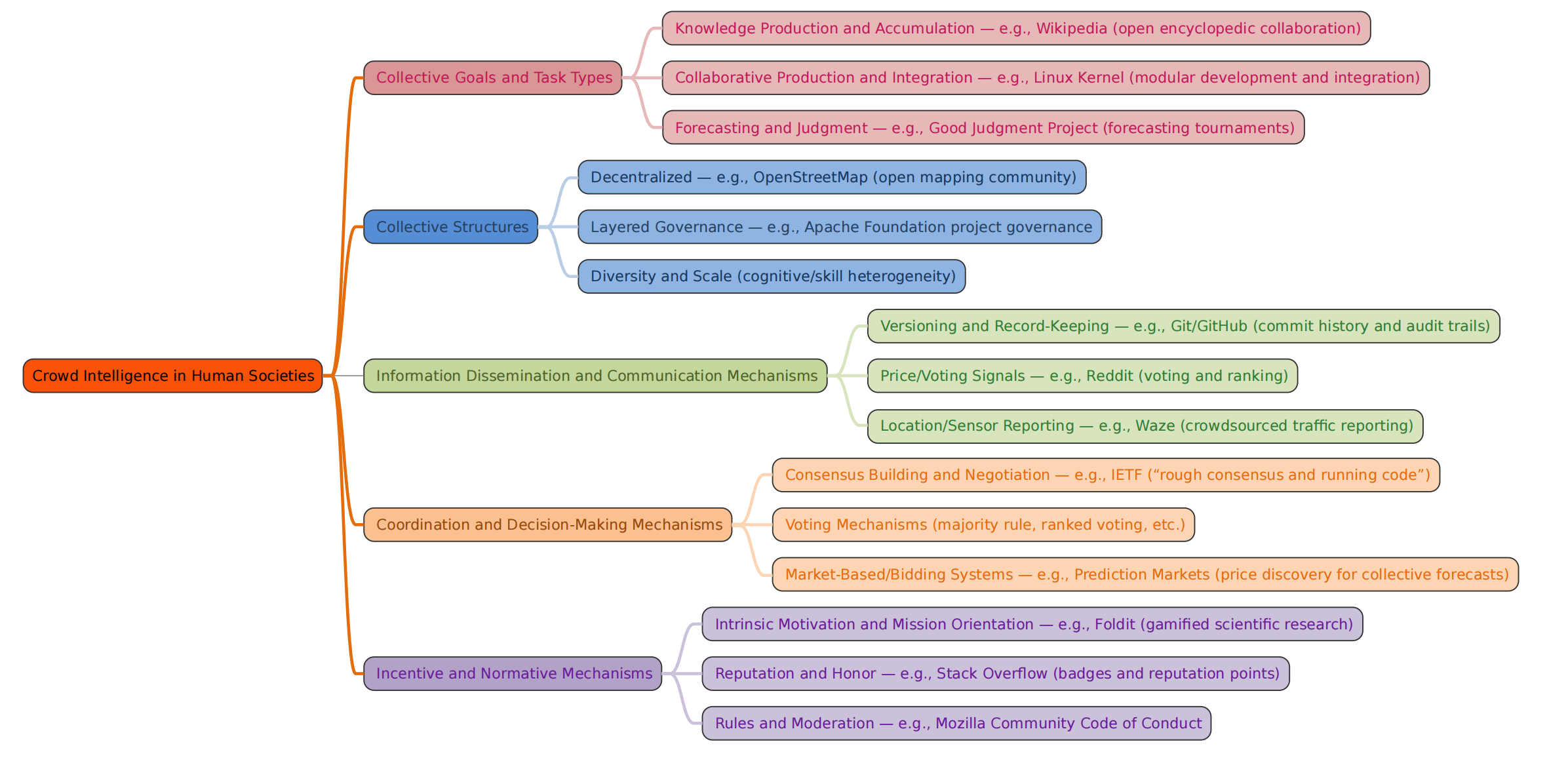}
    \caption{A taxonomy of mechanisms enabling collective intelligence in human societies.}
    \label{fig:sec_3.2_taxonomy}
\end{figure*}

\textbf{Collective intelligence is equally evident in human societies.} At the macro level, millennia of social development and evolution are themselves products of collective intelligence. Most achievements of civilization did not spring from a single genius; they were established gradually through long-term, large-scale collaboration and social production among countless individuals. In both history and current practice, aggregates often outperform individuals. For example, Galton observed in 1907 that the average of hundreds of fairgoers' estimates of an ox's weight was almost exactly correct~\cite{Galton1907}.

In the internet era, large-scale online collaboration makes the manifestation of human collective intelligence even more salient. Networked communication allows thousands of people to interact, share, and cooperate rapidly, producing knowledge and making decisions together. To synthesize research on collective intelligence in human societies, this section organizes the literature around five core dimensions (see Figure~\ref{fig:sec_3.2_taxonomy}):

\begin{itemize} 
\item \textbf{Group goals and task types.} Group goals set the direction for collective action, while task types shape \emph{how—and how readily—}collective intelligence can manifest. Clear, shared goals align individual efforts; vague or contested goals weaken focus and coordination. Task characteristics further determine fit: highly complex tasks that require complementary knowledge and skills suit group collaboration; very simple tasks or those reliant on rare individual insight may suit individuals. 

Decomposability matters: decomposable tasks benefit from division of labor and parallelism; non-decomposable tasks require tight global coordination. From organizational behavior and management, task complexity and variety create scope for diversity, independence, and aggregation ~\cite{Hong2004,Woolley2010}. Evidence from organizational studies shows that strong teams are defined not by size or hierarchy but by psychological safety and shared meaning ~\cite{Edmondson1999}; members feel safe to speak up and drive work from the bottom up.

\item \textbf{Group structure.} Structure determines who participates and how people connect and interact, shaping both production and quality of collective intelligence. Small groups often rely on dense communication; large groups reduce coordination costs through layered or networked designs that range from strict hierarchies to loose, market-like forms. Functional, cognitive, and identity diversity expand the knowledge base, foster innovation, and sharpen collective judgment. Decentralization is a key condition for effective bottom-up collaboration: distributed power and knowledge let individuals act on local information, reducing central bottlenecks and increasing adaptability. By contrast, when structures are closed, power is concentrated and communication is restricted; dissent is suppressed and decision quality declines ~\citep{Tetlock1992}.

\item \textbf{Information diffusion and communication.} Communication mechanisms determine how knowledge flows and integrates across members. Effective communication aggregates dispersed information into collective insight, enabling the group to outperform individuals. Technology greatly widens the scope and depth of interaction. Open collaboration platforms let tens of thousands share documents, exchange views, and correct errors in real time, achieving high coordination without central control. 

Design still matters: overload, redundant diffusion, and herding weaken performance. For example, if discussions revolve only around known information and ignore heterogeneous sources, collective judgment can drift ~\cite{Lorenz2011}. Good mechanism design provides both \textit{channels} and \textit{filters}: it ensures diverse views enter the public space and are transformed into group wisdom through aggregation.

\item \textbf{Coordination and decision-making.} Coordination and decision rules determine how dispersed actions and opinions translate into collective action. Coordination can be centralized, led by core individuals or institutions, suitable for urgent or highly coupled tasks, or decentralized, relying on self-organization and local feedback ~\cite{Malone1994}. Decentralized coordination is common in open communities, prediction markets, and open-source software. It emphasizes consensus, voting, market signals, or algorithmic aggregation to turn diverse opinions into group decisions. Effective mechanisms must aggregate heterogeneous information while preventing authority capture, balancing broad deliberation with timely closure and avoiding domination by a few voices.

\item \textbf{Incentives and norms.} Incentives motivate participation; norms ensure orderly and trustworthy cooperation. Incentives include money, reputation, achievement, interest, and mission. Many open projects rely on non-monetary incentives, especially reputation, to sustain large-scale voluntary effort, whereas prediction markets or bounty platforms rely on financial rewards. Norms align individual behavior with group goals through formal rules, shared values, and informal conventions. Together they shape participation depth, psychological safety, and collective identity, strengthening bottom-up collaborative dynamics. Poorly designed incentives and norms can intensify conformity pressure and superficial consensus, undermining independence and diversity.
\end{itemize}

The following cases illustrate these five dimensions in practice (Table~\ref{tab:sec_3.2_human_cases}).

\begin{table*}[htbp]
\centering
\small
\caption{Comparative analysis of human collective intelligence across five dimensions.}
\label{tab:sec_3.2_human_cases}
\resizebox{\textwidth}{!}{

\begin{tabular}{m{2.5cm} L{3.1cm} L{3.1cm} L{3.1cm} L{3.1cm} L{3.1cm}}
\toprule
\rowcolor{gray!10}
\textbf{Case} & \textbf{Group Goals \& Task Types} & \textbf{Group Structure} & \textbf{Information Diffusion \& Communication} & \textbf{Coordination \& Decision-Making} & \textbf{Incentives \& Norms} \\
\midrule
Wikipedia & Build and continuously update an open encyclopedia; tasks modularized by article. & Decentralized volunteer community; a small number of administrators for governance. & Public revision history and talk pages; transparent, traceable edits. & Consensus first, supplemented by voting and administrator arbitration. & Intrinsic motivation and reputation; adherence to policies and patrolling guidelines. \\
\midrule
Linux open source community & Develop and maintain the OS kernel; modular subsystems evolve in parallel. & Peer based layered structure; maintainers steward subsystems and integration. & Public mailing lists and code hosting records for discussion and review. & ``Propose, review, and merge'' workflow; technical consensus; maintainers decide when needed. & Mix of intrinsic motivation and corporate support; coding standards and code review ensure quality. \\
\midrule
Waze navigation crowdsourcing & Real time aggregation of traffic and map updates; data crowdsourcing and rapid iteration. & Mass users plus tiered volunteer editors co maintain the map. & In app data collection and community forums; real time distribution of traffic information. & Algorithmic route coordination combined with permissioned human moderation. & Gamified points and reputation; editor guidelines and community rules. \\
\midrule
Prediction markets & Aggregate probabilistic judgments on event outcomes via trading contracts. & Open, decentralized trading community; no fixed hierarchy. & Prices immediately reflect information and are visible to all participants. & Market clearing yields collective forecasts; independent trades drive rapid calibration. & Monetary incentives and market rules; anti manipulation safeguards and compliance. \\
\midrule
Foldit citizen science game & Gamified search for optimal protein folds. & Global players with self organization; team collaboration in parallel. & In game real time feedback; community forums share strategies and solutions. & Leaderboards and team merging to combine superior solutions. & Badges and rankings plus scientific mission; anti cheating norms. \\
\bottomrule
\end{tabular}
}
\end{table*}
\vspace{-0.4cm}
\subsection{Collective Intelligence in LLM-based Multi-Agent Systems}
\label{sec: LLM-based Collective Intelligence}
Unlike biological collectives that rely on low-dimensional stimulus–response mechanisms and scale effects, or human societies that depend on institutions and cultural norms, LLM-based collective intelligence is characterized by natural language as the core medium of coordination and language agents—autonomous entities capable of task understanding and reasoning—as the fundamental units. Through linguistic consensus, such systems achieve internal self-organization and collaboration, marking a new stage of explicit symbolization and semanticization in the generation of intelligence~\citep{qian2024chatdev, Park2023, Wu2023}. The following discussion examines its features from three perspectives: the characteristics of interactive agents, their interaction patterns, and their organizational architectures.
\vspace{-0.3cm}
\begin{itemize} 
    \item \textbf{Characteristics of Interactive Agents.} Language agents, powered by large-scale language models, integrate multifaceted abilities such as language understanding, knowledge reasoning, and task execution. They exhibit a high degree of autonomy and adaptability, enabling goal-directed reasoning and self-optimization within complex semantic environments ~\citep{Yao2022, Shinn2023}. Autonomy manifests in their capacities for intent generation, independent decision-making, and self-planning ~\citep{Bran2023}. A language agent can independently interpret task descriptions expressed in natural language, decompose objectives, generate reasoning chains, and form internal representations of execution paths. Adaptability, on the other hand, is reflected in their ability to perform linguistic reflection and strategic adjustment based on environmental feedback and task dynamics, thereby establishing a closed loop of self-improvement ~\citep{Wang2023, Park2023}. 
    Notably, interactive agents' cognitive profiles mirror human-like planning and behavioral adjustment through language-based reasoning. This synergy of autonomy and adaptability defines their intelligent individuality, enabling task comprehension and dynamic optimization—the essential building blocks of LLM-based collective intelligence.
    \item \textbf{Interaction Patterns of Interactive Agents.} Just as humans rely on natural language for communication and collaboration, interactive agents also use natural language as the central medium of coordination, forming the key mechanism of LLM-based collective intelligence. Unlike traditional MAS that communicate through fixed protocols (e.g., FIPA-ACL or KQML) and struggle to handle dynamic contexts or task shifts ~\citep{Labrou1998, Labrou1999}, language agents can dynamically generate, interpret, and adjust semantic structures during interaction, achieving semantic coordination through continual linguistic negotiation ~\citep{Wu2023, li2023camel}. At the collective level, natural language interaction serves not only as a channel for information exchange but also as a driving force for knowledge sharing and group evolution ~\citep{Park2023,Hong2023}. Through continuous semantic negotiation and feedback, interactive agents can gradually form collective memory and shared norms under decentralized conditions, thereby giving rise to stable structures of collaboration. Similar to the cultural role of language in human societies, language-agent communities can achieve a transition from individual understanding to collective cognition through the accumulation of semantic consensus. 
    The interactions among agents—characterized by semantic explicitness, dynamic coordination, and co-construction of knowledge—endow collective intelligence with enhanced flexibility, adaptability, and sociality, forming a language-centered paradigm of ``symbolic collective intelligence.''
    \item \textbf{Organizational Architectures of Interactive Agents.} Analogous to biological collectives, groups of interactive agents display diverse and flexible organizational forms. In nature, wolf packs and lion prides rely on dominant individuals for centralized command and coordination, representing typical centralized architectures ~\citep{Mech1999, Schaller2009}; flocks of birds and schools of fish, conversely, exhibit self-organized global order through local perception and dynamic feedback ~\citep{Couzin2005, reynolds1987flocks}; bee colonies—while centered around the queen—combine symbolic communication mechanisms (e.g., pheromones, waggle dances) with distributed decision-making ~\citep{Seeley2006}, demonstrating a hybrid structure where centralization and decentralization coexist. Similarly, the organizational structures of language-agent systems can be broadly categorized into centralized, decentralized, and hybrid types. In a centralized architecture, a core agent acts as a ``central planner,'' responsible for unified global task management; in a decentralized architecture, agents interact on an equal footing, and group behavior emerges spontaneously from individual interactions; hybrid architectures combine centralized global planning with decentralized local autonomy, maintaining task coherence while preserving agent independence. 
    Selecting or composing an optimal architecture necessitates a strategic trade-off between global control, individual autonomy, and coordination efficiency, tailored specifically to the task’s scale and environmental complexity.
\end{itemize}

\noindent Building upon this foundation, an intriguing question arises: when multiple language-model-based agents form a collective, their interaction patterns become almost indistinguishable from the communication modes of human societies in both form and function. They likewise rely on language as the medium of coordination, achieving shared understanding through semantic negotiation. At the organizational level, they exhibit structural differentiation similar to that observed in human or biological collectives — such as hierarchical role systems, decentralized collaboration networks, and hybrid dynamic organizational forms. This structural and interactive resemblance leads to a profound hypothesis: could collectives of language-model agents also give rise to a form of ``collective intelligence'' analogous to that found in biological or human societies?

\section{Technical Details of Collaboration Gain}
\label{appendix:metric}

\subsection{Task-Specific Baseline Implementation}
\label{app:baseline-std}

We emphasize that the operational instantiation of the baseline $\Phi_S$ is not a fixed constant but a task-dependent ``cognitive ceiling.'' It must be adaptively selected to reflect the intrinsic goals of the task and calibrated to specific resource bottlenecks. Building on these principles, we categorize the aggregation logic of $\Phi_S$ based on the nature of task outputs:

\begin{itemize}
    \item \textbf{Accumulative Tasks (e.g., multi-target tracking):} 
    For tasks where performance is linearly additive, the baseline $\Phi_S$ is defined as the sum of independent contributions derived from dividing the total resource budget among independent single-agent instances. 
    If $\Gamma > 1$, it indicates that collaboration gains—such as dynamic division of labor preventing redundant search paths—have successfully offset the coordination costs, yielding a result superior to simple parallel processing.

    \item \textbf{Coverage Tasks (e.g., mathematical problem sets):} 
    For tasks measured by task set pass rate, the baseline $\Phi_S$ aligns with ``search breadth under equivalent attempts''. Given a total budget of $N$ tokens, $\Phi_S$ is calculated as the union of problems solved by a single agent through independent sampling (e.g., self-consistency with $k$ paths) under that budget. 
    This decouples random gains from ``brute-force sampling,'' ensuring that the $\Gamma$ metric measures the collaborative penetration of complex logic rather than mere probability superposition.

    \item \textbf{Single-Solution Tasks (e.g., code generation, complex planning):} 
    For tasks delivering a single, cohesive output, the baseline $\Phi_S$ represents the ``individual limit after deep deliberation.'' The single agent is granted computational support equivalent to the total MAS budget to execute advanced reasoning strategies (e.g., deeper self-reflection or extended Chain-of-Thought~\citep{Wei0SBIXCLZ22}). 
    This captures whether collective intelligence truly achieves a non-linear leap over the single-agent capability ceiling via interaction, rather than simply thinking longer.
\end{itemize}

\subsection{Dimensions of Resource Equivalence}
\label{app:Resource Equivalence}
To isolate genuine collaboration gain, the MAS and the SAS baseline must be calibrated to the same resource consumption level. Depending on the task bottleneck, ``Resource Equivalence'' refers to one or more of the following dimensions:

\begin{itemize}
    \item \textbf{Model Capability Alignment:} 
    The single-agent baseline must utilize the most capable model base within the MAS. This ensures that the evaluation excludes base-model dividends or illusory gains caused by simply upgrading the underlying foundation model.
    
    \item \textbf{Resource-Matched Evaluation:} 
    The single agent must exhaust a total inference cost (e.g., token length, sampling steps, or tool-invocation counts) equal to that of the entire MAS. This explicitly measures resource consumption to exclude redundant gains derived merely from resource expansion.
\end{itemize}

\section{The MAS Factor Library}
\label{appendix:factor_library}

This appendix provides a literature review and specific implementation for the factor library proposed in section 4. We advocate for a scope that exclusively prioritizes collective-level factors, while deliberately excluding single-agent enhancements—such as memory or tool-use—to ensure a focused diagnostic of collective intelligence. 

{
\small 
\renewcommand{\tabularxcolumn}[1]{m{#1}}

\begin{xltabular}{\textwidth}{
    >{\raggedright\arraybackslash}m{2.2cm}  
    >{\raggedright\arraybackslash}m{2.2cm}  
    >{\raggedright\arraybackslash}m{3.2cm}  
    >{\raggedright\arraybackslash}X       
} 
    \caption{Classification of Optimization Factors for MAS: A Factor Library Based on Prior Work. This table summarizes key optimization factors and their associated high-level strategies from recent MAS studies, offering a structured view of how various approaches impact collective behavior and system performance.} \label{tab:planning_factors_detailed} \\
    
    \toprule
    \rowcolor{gray!10}
    \textbf{Category} & \textbf{Factor} & \textbf{Reference \& Title} & \textbf{Specific Optimization Strategy} \\
    \midrule
    \endfirsthead

    \multicolumn{4}{c}{\textit{(continued from previous page)}} \\
    \toprule
    \rowcolor{gray!10}
    \textbf{Category} & \textbf{Factor} & \textbf{Reference \& Title} & \textbf{Specific Optimization Strategy} \\
    \midrule
    \endhead

    \midrule
    \multicolumn{4}{r}{\textit{(continued on next page)}} \\
    \endfoot

    \bottomrule
    \endlastfoot

    \multirow{8}{=}{\textbf{Task Context (External)}} 
    & \multirow{8}{=}{\textbf{Task Attributes}} 
    & \citep{Fan2025} \newline \textit{MM-Agent} 
    & \textbf{Sequential Dependency:} Decompose open-ended tasks into four distinct sequential stages. \\
    \cmidrule{3-4}
    & 
    & \citep{bigeard2025finance, kim2025towards} \newline \textit{Finance Agent}
    & \textbf{High Task Decomposability:} Enable sub-tasks to be executed in parallel through a centralized coordinator. \\
    \cmidrule{3-4}
   & 
     & \citep{tang2025on} \newline \textit{Depth-Width Writing ($DW^2$)}
     & \textbf{Task Complexity:} Model tasks by Depth (sequential reasoning steps) and Width (capability diversity). Prioritize MAS for high-depth tasks. \\
      \cmidrule{3-4}
      &
     & \citep{huotagents} \newline \textit{Agents' Room} 
     & \textbf{Task Openness:} Open-ended tasks (e.g., fiction writing) inherently favor MAS through specialized sub-task decomposition. \\

    \midrule 

    \multirow{16}{=}{\textbf{MAS Construction (Internal)}} 
  
    & \multirow{4}{=}{\textbf{Agent Scale}}
    & \citep{qian2024scaling} \newline \textit{MacNet} 
    & \textbf{Topological Scaling via DAGs:} Structure massive agent populations into Directed Acyclic Graphs (DAGs). \\
    \cmidrule{3-4}
    & 
    & \citep{Wang2025} \newline \textit{MegaAgent}
    & \textbf{SOP-Free Dynamic Scaling:} Dynamically generate agent populations to match real-time task complexity. \\
    \cmidrule{2-4}

    & \multirow{6}{=}{\textbf{Agent Diversity}}
    & \citep{Rui2025} \newline \textit{X-MAS}
    & \textbf{Heterogeneous Model Integration:} Leverage model heterogeneity by assigning specialized LLMs to agent roles based on domain-specific capabilities. \\
    \cmidrule{3-4} 
    & 
    & \citep{Dongfu2023} \newline \textit{LLM-Blender}
    & \textbf{Ensemble-Based Generative Fusion:} Aggregate diverse model outputs via pairwise ranking and generative fusion to synthesize a superior response. \\
    \cmidrule{3-4} 
    & 
    & \citep{Wang20241} \newline \textit{MoA}
    & \textbf{Layered Collaborative Refinement:} Orchestrate diverse LLMs in a multi-layer structure where agents iteratively refine responses by attending to collective outputs from the previous layer. \\
    \cmidrule{2-4}

    & \multirow{4}{=}{\textbf{Organizational Structure}}
    & \citep{Chen20251} \newline \textit{Puppeteer}
    & \textbf{RL-Driven Dynamic Orchestration:} Dynamically sequence agent interactions via an RL-trained orchestrator that adapts to evolving task states without static workflows. \\
    \cmidrule{3-4}
    & 
    & \citep{Chan2023} \newline \textit{ChatEval}
    & \textbf{Multi-Agent Debate Topology:} Organize agents into a debate structure to synthesize diverse perspectives through autonomous multi-round argumentation. \\
     \cmidrule{3-4}
    & 
    & \citep{gong2024mindagent} \newline \textit{MindAgent}
    & \textbf{Centralized Task Coordination:} Dynamically schedule tasks and dispatch instructions to coordinate concurrent multi-agent and human interactions. \\
     \cmidrule{3-4}
     &
    & \citep{Zhou2025} \newline \textit{MASS}
    & \textbf{Interleaved Design Optimization:}Iteratively optimize local prompts, workflow topologies, and global prompts in a multi-stage search process. \\
     \cmidrule{3-4}
    & 
    & \citep{Zhang20251} \newline \textit{MaAS}
    & \textbf{Supernet-Based Architecture Search:} Dynamically sample query-dependent architectures from a probabilistic supernet. \\
    \cmidrule{2-4}
    
    &\multirow{12}{=}{\textbf{Communication Mechanism}}
    & \citep{Zhang2024} \newline \textit{AgentPrune} 
   & \textbf{One-Shot Graph Pruning:} Use one-shot pruning on the spatial-temporal message-passing graph to eliminate redundant interactions. \\
    \cmidrule{3-4}
    
    & 
    & \citep{Yang2025} \newline \textit{Entropy-Debate} 
   & \textbf{Entropy Compression \& Adversarial Debate:} Enhance communication efficiency via entropy-compression and employ adversarial debates and voting mechanisms for cross-verification. \\
    \cmidrule{3-4}
    
    & 
    & \citep{Chen2025} \newline \textit{Optima} 
    & \textbf{Token-Efficient Training:} Leverage SFT/DPO with cost-aware rewards to balance task performance against token consumption in agent dialogues. \\ 
    \cmidrule{3-4}

     &
    & \citep{Zou2025} \newline \textit{LatentMAS} 
    & \textbf{Training-Free Latent Collaboration:} Establish a shared latent working memory to transfer latent thoughts (last-layer hidden embeddings) directly. \\
   
\end{xltabular}
}

\section{Validation of the Factor Attribution Paradigm}
\label{appendix:feasibility_validation}
This appendix validates the factor attribution paradigm proposed in the main text, specifically demonstrating the utility of collaboration gain ($\Gamma$) as a diagnostic probe. As an exploratory pre-experiment to assess the feasibility of $\Gamma$ as a probe for control-level factors, we systematically varied specific factors in the factor library (organization structure, agent diversity, agent scale) and examined the corresponding changes in $\Gamma$. Our observations in this specific setting suggest that this metric can help distinguish between genuine collaboration gain ($\Gamma>1$) and resource redundancy ($\Gamma<1$).

\subsection{Validation I: Diagnostic Sensitivity of $\Gamma$ to Factors}
\label{appendix:sensitivity_validation}

To validate the efficacy of $\Gamma$ as an effective probe for control-level factors, a preliminary sensitivity analysis was first conducted, with a specific focus on organizational structure. We measured $\Gamma$ across three distinct datasets (SRDD~\citep{qian2024chatdev}, IdeaBench~\citep{ideabench}, and CommonGen\citep{lin2020commongen}).

\begin{figure}[H]
    \centering
    \includegraphics[width=0.5\linewidth]{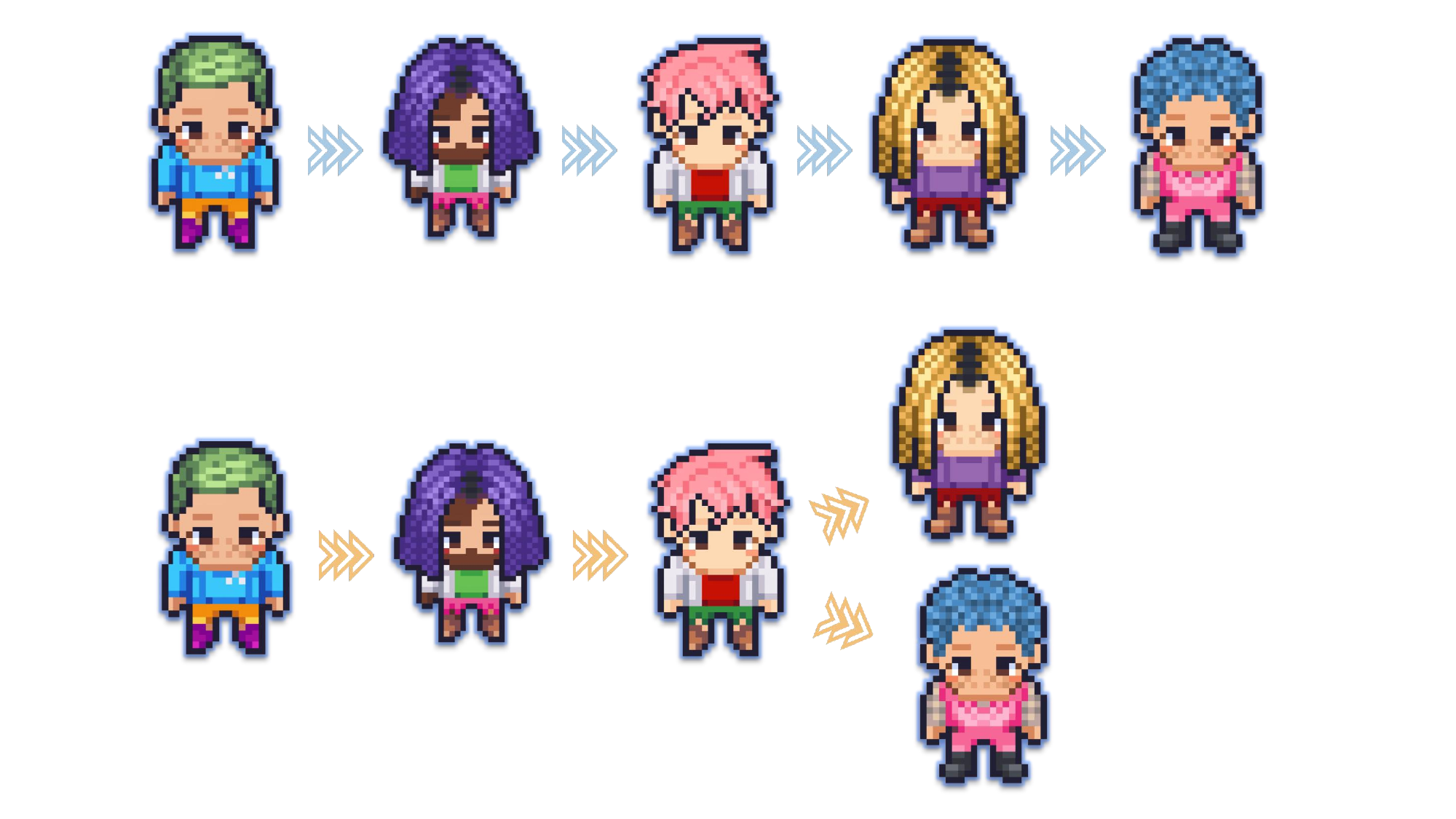}
    \caption{Visual Representation of Agent Topological Structures.
    The upper row illustrates the chain topology, characterized by a sequential, unidirectional data flow among five distinct agents. The lower row depicts a tree topology (specifically a branching structure), where the workflow diverges after the third agent.}
    \label{fig:agent_topology_visual}
\end{figure}

We held agent scale (5 Agents) and model backbone (Qwen3-30B-A3B-Instruct-2507) constant while varying only the organizational-structure configuration between a sequential chain and a branching tree. For each benchmark, we randomly sampled 30 distinct tasks to evaluate the system's performance, reporting the averaged metrics.

For baseline construction, the SAS serves as a resource-equivalent reference, enhanced with extended CoT within the same total token budget as the MAS. This ensures that $\Gamma > 1$ identifies genuine structural dividends rather than advantages derived from allowing a single agent to "think longer" under the same resource allocation.

\begin{figure}[H]
    \centering
    \includegraphics[width=0.5\linewidth]{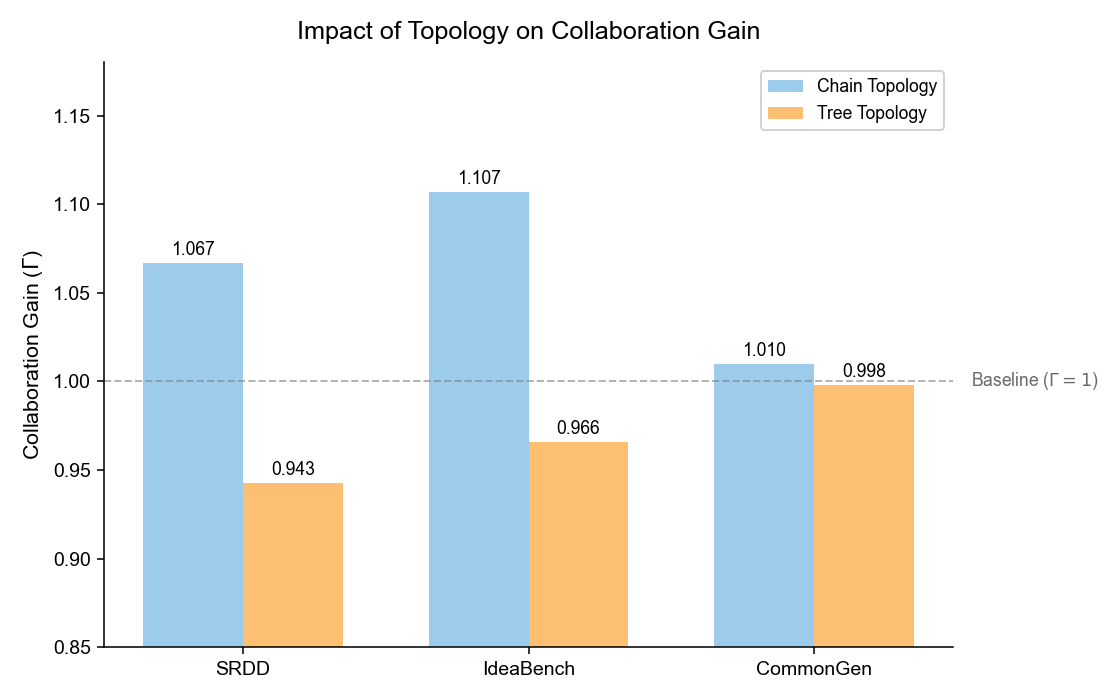}
    \caption{Impact of Topological Structure on Collaboration Gain ($\Gamma$)}
    \label{fig:gamma_topology}
\end{figure}
\vspace{-0.5cm}

The results in Figure~\ref{fig:gamma_topology} suggest the metric's potential sensitivity in distinguishing topological effectiveness within the tested domains. Specifically, observations of $\Gamma>1$ in the chain topology suggest that this structure facilitated information accumulation in these tasks; however, the modest magnitude of these gains indicates that this topology is not a universal remedy, as it remains vulnerable to collapse under naive scaling strategies (detailed in the~\ref{sec:phaseIII} analysis). In contrast, the metric reflected the performance penalty in tree topologies ($\Gamma<1$) for these specific scenarios, indicating where the overhead of coordinating branching paths might negate the potential benefits of parallelization.

In this specific experimental setting, the metric $\Gamma$ accurately reflected the performance disparity between the two topologies. It captured the performance drop observed in the tree topology ($\Gamma < 1$) and the gain in the chain topology ($\Gamma > 1$). Within the context of chain and tree topologies evaluated here, this demonstrates the potential of $\Gamma$ as a diagnostic signal to detect effective versus reduced collaboration outcomes within a given context, serving as a basis for the subsequent factor analysis.

\subsection{Validation II: Factor Attribution Mechanism: A detailed case study}
\label{appendix:detailed_case}

Having established the general sensitivity of the collaboration gain metric ($\Gamma$), we present a detailed case study of the ``Navigation Buddy'' task from the SRDD dataset~\citep{qian2024chatdev}.Our objective is not to propose a state-of-the-art multi-agent framework, but to utilize this case as a controlled environment to illustrate how the collaboration gain metric might assist in factor attribution and inform the evolutionary construction of MAS. We analyze three phases through step-wise factor modification.

\subsubsection{Experimental Setup and Controls}
\label{sec:exp_setup}

\textbf{Experimental Control and Budget.} To eliminate computational volume as a confounding factor, we enforce a strict token budget ($B_{max} = 20,000$) for both systems. The MAS distributes this budget across all agents and interaction rounds, while the SAS is allocated the full equivalent budget. Crucially, the SAS is explicitly enhanced with CoT strategies to ensure a ``strong baseline,'' ensuring that performance gaps are attributed to architectural intelligence rather than computational quantity.

\textbf{Evaluation Metrics.} We adopt objective metrics from ChatDev~\citep{qian2024chatdev}, focusing on completeness, executability, and consistency. We define software quality ($Q$) as the product of these three dimensions ($Q = \text{completeness} \times \text{executability} \times \text{consistency}$). This composite metric is utilized to calculate the collaboration gain ($\Gamma$), providing a unified standard for assessing software reliability and adherence to requirements.

\textbf{Model Consistency.} To ensure fairness, all employed models belong to the same 30B parameter class. We utilize \textit{Qwen3-30B-A3B-Instruct-2507} as the foundation model for the SAS and all MAS roles in phase I.

\subsubsection{Phase I: Probing the ``Agent Diversity(Role Diversity)'' Factor: $\Gamma > 1$}

\textbf{Factor Configuration Table}

\begin{longtable}{lp{9cm}}
\caption{MAS Construction (Internal) based on section 4 Framework} \label{tab:mas_construction} \\
\toprule
Factor & Configuration \\
\midrule 
Organization Structure & Chain \\
Communication Mechanism & Natural Language\\
Agent Diversity & Role Diversity (Heterogeneous Roles: CEO for Strategy, CTO for Architecture, Programmer for Coding). \\
Agent Scale & 3 Agents (1 CEO, 1 CTO, 1 Programmer) \\
\bottomrule
\end{longtable}

\textbf{Qualitative Analysis}

We conducted a qualitative analysis of the multi-agent collaboration by examining the logged interaction records. Due to space constraints, we present only selected outputs that highlight the collaborative advantage of the MAS. Each retained output is annotated to indicate its strategic significance.

\begin{tcolorbox}[
    colback=role_ceo!30!white,
    colframe=role_ceo!80!black,
    title=\textbf{User - CEO: Core CEO Output (Selected)},
    breakable,
    fonttitle=\bfseries
]

Define operational mechanics, workflow, and quality constraints for Navigation Buddy, a real-time navigation system.

Ensure developers implement robust, shortcut-free code; maintain production reliability and security.

\textcolor{blue}{Analysis: Centralizes strategy and standards, reducing risk of fragmented or inconsistent implementation.}

\end{tcolorbox}

\begin{tcolorbox}[
    colback=role_cto!30!white,
    colframe=role_cto!80!black,
    title=\textbf{CEO - CTO: Core CTO Output (Selected)},
    breakable,
    fonttitle=\bfseries
]

\begin{lstlisting}[language=json, caption={Project File Structure}]
{
  "file_structure": [
    "navigation_buddy/__init__.py",
    "navigation_buddy/config.py",
    "navigation_buddy/core/__init__.py",
    "navigation_buddy/core/position.py",
    "navigation_buddy/core/route_planner.py",
    "navigation_buddy/core/traffic_data.py",
    "navigation_buddy/integrations/gps_provider.py",
    "navigation_buddy/integrations/map_provider.py",
    "navigation_buddy/services/__init__.py",
    "navigation_buddy/services/directions_service.py",
    "navigation_buddy/services/realtime_updates.py",
    "navigation_buddy/utils/__init__.py",
    "navigation_buddy/utils/logging.py",
    "navigation_buddy/utils/time_utils.py",
    "navigation_buddy/main.py"
  ]
}
\end{lstlisting}

\textcolor{blue}{Analysis: Transform an open-ended problem into a constrained design space.}

\end{tcolorbox}

\begin{tcolorbox}[
    colback=role_dev!30!white, 
    colframe=role_dev!80!black, 
    title=\textbf{CTO - Programmer: Core Programmer Output (Selected)},
    breakable,
    fonttitle=\bfseries
]

\begin{lstlisting}[language=Python, caption={Input Validation \& Logic Orchestration}]
    def get_directions(self, start, end):
        # CONSTRAINT CHECK: Input Validation
        if not self.location_service.validate_location(start):
            raise ValueError("Invalid start location")   
        if not self.location_service.validate_location(end):
            raise ValueError("Invalid end location")
        # Dynamic Resolution
        if start == "Current Location":
            start = self.location_service.get_current_location()
        # Core Business Logic
        route = self.routing_engine.compute_route(start, end)
        return {
            "start": start,
            "end": end,
            "route": route,
            "steps": self._format_steps(route)
        }
\end{lstlisting}

\textcolor{blue}{Analysis: This reflects CTO-defined architectural contracts, reducing coupling and error propagation paths.}
\end{tcolorbox}

\begin{tcolorbox}[
    colback=result_fail!10!white, 
    colframe=result_fail!80!black, 
    title=\textbf{Single Agent System (SAS): Evidence Output (Selected)},
    breakable,
    fonttitle=\bfseries
]
\begin{lstlisting}[language=Python, caption={Stubbed \& Linear Logic}]
def get_current_location():
    # Returns constant instead of implementation
    return "XYZ" 

def navigate_to(location):
    # Linear script ignoring required event loops
    route = calculate_route(current, location)
    print(route)
\end{lstlisting}

\end{tcolorbox}

\vspace{0.5em}

We compare the generated code from the multi-agent system against a resource-equivalent single-agent baseline. Despite CoT prompting, the SAS showed consistency issues in long-generation tasks, forgetting planning-phase constraints and using placeholders as the context window filled.

\textbf{Quantitative Evaluation}

We utilize the composite $Q$ score as the primary basis for calculating the collaboration gain ($\Gamma$) proposed in Section~\ref{sec:genuine-collaboration-gain}.
We compare the MAS against the resource-equivalent SAS. In this selected case study, both systems can successful execute; however, disparities persisted in completeness and consistency.
\begin{table}[H]
\centering
\caption{Quantitative comparison of code generation performance} \label{tab:gain_analysis}
\begin{tabular}{lccc|c|c}
\toprule
Setting & Comp. & Exec. & Cons. & \textbf{$Q$} & \textbf{$\Gamma$}\\
\midrule
Single Agent (Baseline) & 0.35 & 1 & 0.76 & 0.27 & / \\
Multi-Agent System & \textbf{0.42} & \textbf{1} & \textbf{0.81} & \textbf{0.34} & \textbf{1.26} \\
\bottomrule
\end{tabular}
\end{table}
The MAS achieved a collaboration gain ($\Gamma > 1$), demonstrating better performance than the resource-equivalent single-agent baseline.

\subsubsection{Phase II: Optimizing via ``Agent Diversity (Model)'' Factor ($\Gamma \uparrow$)}

Building upon the findings from the phase I experiment, where functional orthogonality via role diversity resulted in a positive collaboration gain, we posit that the \textit{agent diversity} factor can be further optimized through Model Heterogeneity.

\textbf{Factor Configuration Update}

 In phase II, we introduce model heterogeneity by switching the programmer agent to \textit{Qwen3-Coder-30B-A3B-Instruct}. This setup maintains a constant parameter scale while isolating the impact of specialized coding expertise.

\textbf{Qualitative Analysis}

Rather than increasing surface complexity, the change manifested as the emergence of new engineering capability classes that were absent in the homogeneous configuration.
\begin{tcolorbox}[
  colback=role_dev!15!white,
  colframe=role_dev!80!black,
  title=\textbf{New Programmer Output (Selected)},
  fonttitle=\bfseries,
  breakable
]

\begin{lstlisting}[language=Python]
class RealTimeUpdates:
    """Manages real-time navigation updates"""
    def start_monitoring(self):
        if not self.is_monitoring:
            self.is_monitoring = True
            # Autonomous decision: Threaded execution for non-blocking I/O
            self.monitoring_thread = threading.Thread(
                target=self._monitor_loop
            )
            self.monitoring_thread.daemon = True
            self.monitoring_thread.start()
    def _monitor_loop(self):
        while self.is_monitoring:
            try:
                self._process_updates()
                time.sleep(30)
            except Exception as e:
                print(f"Error in monitoring loop: {e}")
\end{lstlisting}

\textcolor{blue}{Analysis: An autonomous concurrency model that was not specified at the architectural level.}
\end{tcolorbox}

The output demonstrates enhanced structural completeness, exemplified by features such as threaded concurrency and observability decorators.

\textbf{Quantitative Evaluation}

To rigorously quantify the effect of the enhanced agent diversity factor, we revisit the collaboration gain ($\Gamma$) metric based on the composite software quality score ($Q$). Table~\ref{tab:gain_analysis_comparison} reports the comparative results. The data exhibits a stepwise improvement in system performance as the agent diversity factor is progressively enriched.

\begin{table}[h]
\centering
\caption{Quantitative comparison of code generation performance}
\label{tab:gain_analysis_comparison}
\begin{tabular}{lccc|c|c}
\toprule
Setting & Comp. & Exec. & Cons. & \textbf{$Q$} & \textbf{$\Gamma$} \\ 
\midrule
Single Agent (Baseline) & 0.35 & 1 & 0.76 & 0.27 & / \\  
Multi-Agent System & \textbf{0.60} & \textbf{1} & \textbf{0.83} & \textbf{0.50} & \textbf{1.85} \\ 
\bottomrule
\end{tabular}
\end{table}

The increase in $\Gamma$ corresponds to the introduction of model heterogeneity. In this case, this factor appeared to facilitate a capability transition from ``functional correctness'' to ``engineering abstraction'' (e.g., the autonomous implementation of concurrency patterns), suggesting that strategic model-role alignment can contribute to system performance.

\subsubsection{Phase III: Boundary Testing the ``Agent Scale'' Factor ($\Gamma < 1$)}
\label{phase:3}\label{sec:phaseIII}
Following the validation of role heterogeneity (phase I) and model heterogeneity (phase II), we examine a deeper scalability question: Does change the factor of agent scale produce better collaboration gain? To find and verify this factor hypothesis, we assume that expanding the implementation layer—by stacking multiple Programmer agents in sequence—should improve output quality through iterative refinement, redundancy, and error correction.

\textbf{Factor Configuration Update}

To isolate the effect of agent scale, the experimental configuration was updated to a sequential implementation chain expanded to 5 agents. To ensure valid collaboration within this sequential topology, Programmer 2 and 3 were explicitly instructed to ingest the predecessor's code and perform incremental optimization (e.g., bug fixing and logic hardening).

\textbf{Qualitative Analysis}  

Contrary to expectations, the qualitative analysis indicates a decline in the multi-agent system's performance in this sequential setup. In this sequential setup, we observed a structural regression in the multi-agent system's output. Following the initial complexity construction by Programmer 1, the subsequent agent did not maintain the multi-file architecture. Instead, the final output was simplified into a single, self-contained script, prioritizing executability over the original structural integrity.

\begin{tcolorbox}[
colback=result_fail!10!white,
colframe=result_fail!80!black,
title=\textbf{Interaction Output(Selected)},
fonttitle=\bfseries,
breakable
]

\begin{lstlisting}[language=Python]
# main.py (Programmer 2)
class NavigationBuddy:
    def __init__(self):
        self.current_location = (0.0, 0.0)
        self.route = [
            "Head east for 2.5 km...",
            "Turn right onto Main Street."
        ]
def main():
    lat_input = float(input("Latitude: "))
\end{lstlisting}

\textcolor{red}{\textbf{Analysis:} The agent fails to preserve the multi-file architectural model. To maintain executability, it collapses the system into a monolithic script.}

\end{tcolorbox}

\textbf{Quantitative Evaluation}

\begin{table}[H]
\centering
\caption{Quantitative comparison of code generation performance}
\label{tab:gain_analysis_phase3}
\begin{tabular}{lccc|c|c}
\toprule
Setting & Comp. & Exec. & Cons. & \textbf{$Q$} & \textbf{$\Gamma$} \\ 
\midrule
Single Agent (Baseline) & 0.35 & 1 & 0.76 & 0.27 & / \\
Multi-Agent System & \textbf{0.23} & \textbf{1} & \textbf{0.74} & \textbf{0.17} & \textbf{0.63} \\
\bottomrule
\end{tabular}
\end{table}

Collaboration gain collapses to $\Gamma<1$, indicating that, within this specific experimental design, naive scale expansion did not yield cumulative intelligence.

\textbf{Process Dynamics: Diagnostic Analysis via Information Flow}

\begin{figure}[H]
\centering
\includegraphics[width=0.5\linewidth]{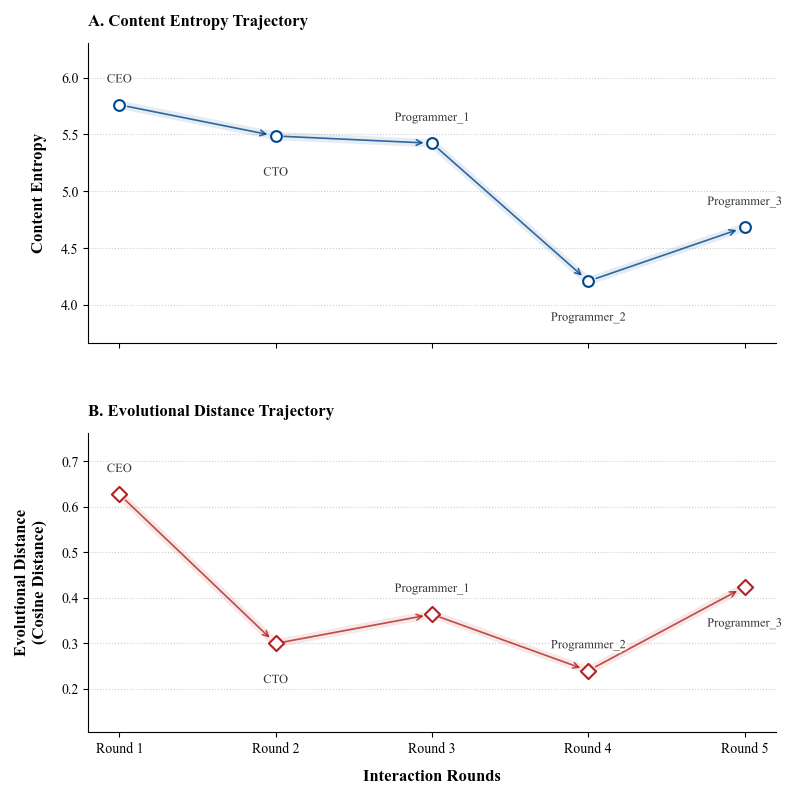}
\caption{Evolutionary Dynamics of Information Level}
\label{fig:process_dynamics}
\end{figure}

\vspace{-0.5cm}

To interpret the collapse in collaboration gain ($\Gamma<1$), we conducted a diagnostic analysis combining the detailed interaction logs with the information flow metrics shown in Figure \ref{fig:process_dynamics}. 

In the initial phase, the system exhibits a stable transmission pattern: content entropy shows a gradual decline, while evolutionary distance increases (particularly from CTO to Programmer\_1). This trajectory corresponds to a healthy ``semantic expansion,'' where abstract requirements are effectively translated into concrete code implementations without losing focus. However, a critical anomaly emerges during the transition from Programmer\_1 to Programmer\_2. Here, we observe a precipitous drop in both content entropy and evolutionary distance. Cross-referencing this with the interaction logs suggests a ``fracture'' in the information flow caused by context overload. The extensive multi-file architecture generated by Programmer\_1 appears to have exceeded the effective processing window of the subsequent agent, causing Programmer\_2 to discard structural details and focus on a narrowed subset of the data. Finally, the subsequent rise in metrics at Programmer\_3 indicates that the agent generated new content, but this expansion was grounded in the already broken, incomplete context provided by Programmer\_2. This fluctuation implies that the observed loss of collaboration gain likely stems from a bottleneck in sustaining complex context across extended sequential chains.

\subsection{Conclusion}
The analysis presented above, preliminarily illustrates that collaboration gain  serves as a principled diagnostic signal for factor attribution within MAS. By controlling for resource consumption, $\Gamma$ enables the identification of specific factors—such as agent diversity—that either facilitate or impede system performance. Combined with information-level metrics, $\Gamma$ provides feedback for detecting internal bottlenecks, offering guidance for evaluating and refining MAS beyond empirical observation.

\end{document}